\ifdefined\synctex\synctex=1\fi
\documentclass[12pt]{article}
\usepackage[utf8]{inputenc}
\usepackage[english]{babel}
\usepackage[margin=1in]{geometry}
\usepackage{amsmath,amssymb,amsthm,mathtools,bm}
\usepackage{float}
\usepackage{algorithm,algorithmic}
\algsetup{linenodelimiter=.,linenosize=\normalsize}

\usepackage[round]{natbib}
\usepackage{xurl}
\usepackage{hyperref}
\PassOptionsToPackage{nopatch=footnote}{microtype}
\usepackage{bbm,enumitem,microtype,etoolbox}
\AtBeginEnvironment{thebibliography}{\interlinepenalty=10000}
\newtheorem{theorem}{Theorem}[section]

\newtheorem{lemma}[theorem]{Lemma}
\newtheorem{corollary}[theorem]{Corollary}
\theoremstyle{definition}
\newtheorem{assumption}[theorem]{Assumption}

\theoremstyle{remark}
\newtheorem{remark}{Remark}[section]

\newcommand{\KL}{\operatorname{KL}}
\newcommand{\TV}{\operatorname{TV}}

\newcommand{\essinf}{\operatorname*{ess\,inf}}

\DeclareMathOperator{\Unif}{Unif}

\hypersetup{colorlinks=true,linkcolor=blue,citecolor=blue,urlcolor=blue,
pdftitle={Robust Actor-Critic Learning under Transport and Divergence Uncertainty},
pdfauthor={Lican Kang},pdfsubject={Theoretical research manuscript},
pdfkeywords={Robust Markov decision process, actor-critic, distributional robustness, generative policy, neural estimation}}

\author{
Kunqi Huang
\thanks{School of Mathematics and Statistics,
Wuhan University, Wuhan, 430072, China. Email: huangkunqi@whu.edu.cn }
\and
Lican Kang
\thanks{Institute for Math and AI,   Hubei Key Laboratory of Computational Science, and
School of Artificial Intelligence,
Wuhan University, Wuhan, 430072, China. Email: kanglican@whu.edu.cn }
\and 
Hongrui Shu
\thanks
{School of Artificial Intelligence,
Wuhan University, Wuhan, 430072, China. Email: shuhongrui18@gmail.com }
\and
Jincheng Ying 
\thanks{
School of Mathematics and Statistics,
 Wuhan University, Wuhan, 430072, China.
Email: jinchengying@whu.edu.cn
}
}
\date{}

\title{Robust Actor--Critic Learning under Transport and Divergence Uncertainty}
\begin{document}
\clubpenalty=10000
\widowpenalty=10000
\maketitle

\begin{abstract}
Learning reliable control policies requires accounting for uncertainty
in transition dynamics and errors introduced by function
estimation and action generation. We develop an actor--critic method
for discounted robust Markov decision processes with continuous states
and actions and an unknown nominal transition law. We consider transition
uncertainty sets defined by Wasserstein distance, total variation,
Kullback--Leibler divergence, or Pearson chi-square divergence. A
conditional diffusion model is fitted once to offline transitions.
At each subsequent iteration, we estimate the robust action-value
function from current-policy observations using a shared-target minimax
criterion, then update a generative actor through exponentially weighted
score matching.
Regularization preserves exploration and provides the density
regularity needed for conditional distribution estimation. Under
coverage and regularity conditions, we establish finite-sample bounds
for  critic estimation, the learned conditional action law, and
the robust return of the implemented policy. The bounds distinguish
critic estimation error, uncertainty in the nominal transition model,
and policy regularization, and characterize how the four robustness
radii affect the resulting guarantees.
\end{abstract}


\section{Introduction}\label{sec:intro}
Reinforcement learning (RL) uses observed transitions to learn
sequential decision rules. The resulting policy is evaluated through
its long-run rewards, but its performance also depends on whether the
training dynamics represent those encountered at deployment. A robust Markov
decision process (MDP) makes this dependence explicit by evaluating
policies against a prescribed set of transition laws. State--action
rectangular uncertainty permits a conditional worst-case expectation at
each decision and leads to robust dynamic programming
\citep{Iyengar2005,NilimElGhaoui2005}. When the nominal dynamics are
unknown, the uncertainty set is centered on a model learned from data.
The resulting statistical error is distinct from the deliberate
protection supplied by the robustness radius: both influence policy
performance and must be accounted for.
This distinction matters in continuous control, where transition laws
and action values cannot generally be tabulated. Finite-sample
guarantees are available for tabular robust learning with nominal generative
access \citep{XuPanagantiKalathil2023}, offline robust fitted iteration
with function approximation \citep{Panaganti2022RFQI}, and continuous
model-based learning under structured unknown dynamics
\citep{Ramesh2024}. A common difficulty is that robust policy evaluation
contains a nonlinear conditional worst-case expectation. Wasserstein distance,
total variation (TV), Kullback--Leibler (KL) divergence, and Pearson
$\chi^2$ divergence produce different dual problems and different
sensitivities to errors in their nominal centers
\citep{GaoKleywegt2023,DuchiNamkoong2019,XuPanagantiKalathil2023}.
Replacing the nominal law by an empirical or fitted distribution
therefore introduces a bias that ordinary conditional regression
does not remove. A policy-evaluation criterion must account separately for
this bias and finite-sample estimation error.

Actor--critic learning couples this evaluation problem with repeated
policy improvement. Robust natural actor--critic (RNAC) provides finite-time
guarantees with function approximation \citep{Zhou2023}; robust mirror
descent supplies relative policy updates \citep{LiLanZhao2022,wangrobust2024}.
Distributionally robust soft actor--critic (DR-SAC) further combines
learned nominal dynamics with KL-robust,
entropy-regularized actor--critic learning \citep{drsac2026}.
For continuous actions, a generative actor can represent the relative
exponential improvement target without evaluating its normalizing
constant. Exponentially reweighted denoising implements this
target from current-policy samples \citep{Ma2025Diffusion}.
A remaining statistical challenge is to connect the actor implemented
with finite sampling, rather than its ideal target, to subsequent robust
policy evaluation and the final policy's unregularized robust return. Conditional distribution-estimation theory provides a basis for this
analysis, but it must be applied to the actual weighted target
and combined with robust target and policy-regularization errors.

In this work, we develop a robust actor--critic method on continuous
state and action cubes for the four transition-uncertainty families.
A nominal conditional diffusion model (CDM) is fitted once to offline transitions
and then held fixed. Each iteration evaluates the currently implemented
policy through a shared-target minimax criterion and learns a
regularized exponential policy improvement from weighted current-policy
samples.

\subsection{Contributions}
\begin{itemize}
\item  We develop a robust actor--critic method for four
rectangular transition-uncertainty families using one fixed nominal
conditional diffusion model. Conditional duality supplies robust Bellman
targets, and a shared-target minimax criterion estimates the current
policy's action value. Exponentially weighted score learning implements
relative policy improvement without evaluating policy densities or target
normalizing constants. Gaussian smoothing and uniform exploration provide
the regularity required for repeated evaluation of the implemented
generative actor.
\item   We establish finite-sample bounds for on-policy $L^2$
critic error, conditional action-distribution error, and last-iterate robust
return. ReLU DNN approximation and conditional diffusion theory control
critic and actor estimation, while retaining nonlinear target bias,
nominal-model error, and the distinct effects of the four robustness radii.
Robust Bellman contraction connects these errors to policy performance,
with a joint allocation of critic observations, offline transitions, and
samples for critic integration and actor fitting.
\end{itemize}

\subsection{Related work}
\paragraph{Robust learning.}
Rectangular transition uncertainty supports dynamic programming
\citep{Iyengar2005,NilimElGhaoui2005}. Finite-sample robust learning extends beyond this planning formulation:
\citet{XuPanagantiKalathil2023} analyze the same four uncertainty
families in tabular episodic problems with nominal generative access,
while \citet{Panaganti2022RFQI} develop offline TV-robust fitted
$Q$-iteration with function approximation.
The latter guarantees retain completeness, dual approximation,
and coverage requirements.
For continuous dynamics, \citet{Ramesh2024} combine active
Gaussian-process model learning with TV, KL, and $\chi^2$ robustness.
RNAC uses nominal on-policy observations with double-sampling or
integral-probability-metric uncertainty and analyzes both linear
and general function approximation \citep{Zhou2023}.
More recently, online general-class guarantees have been expressed
through robust Bellman--Eluder complexity \citep{Ghosh2026}.

\paragraph{Value estimation.}
Neural fitted-value and policy-iteration methods connect regression
error to policy performance \citep{feng2023,jiao2025}.
The shared-target criterion used in Schr{\"o}dinger--F{\"o}llmer
actor--critic (SFAC) cancels a target's conditional
variance through the difference of two squared losses \citep{sfac2026}. With nonlinear
robust targets, the conditional-mean bias remains, and the auxiliary
class must approximate the Bellman images of all candidate critics.
Our ReLU DNN classes verify this approximation requirement without
imposing unrestricted Bellman completeness.
Robust control and robust inference address different targets.
\citet{bennett2024} derive sharp fixed-policy bounds under
multiplicative transition perturbations and construct orthogonal,
semiparametrically efficient estimators. Here, the policy changes
during learning, and the uncertainty is specified by the four
transition neighborhoods.
DR-SAC is a closer algorithmic comparison: it learns nominal
generative dynamics for offline KL-robust soft actor--critic
\citep{drsac2026}. Its objective includes policy entropy.
Our analysis instead targets unregularized robust return and retains
the empirical-center bias and the law induced by finite sampling from
the learned actor.

\paragraph{Generative policies.}
Denoising score matching estimates distributional scores using Gaussian
perturbations \citep{Vincent2011}. Diffusion-QL combines a conditional
diffusion policy with value-guided behavior cloning in offline learning
\citep{Wang2023Diffusion}, while Q-score matching relates policy scores to
action-value gradients in off-policy learning \citep{Psenka2024}.
Q-weighted variational policy optimization (QVPO) weights denoising
losses by transformed action values
\citep{Ding2024}. Closest to our actor update, \citet{Ma2025Diffusion}
derive reweighted score matching for the relative exponential policy
target and express its loss through exponentially weighted samples from
the current policy. SFAC realizes a relative tilt through a
Schr{\"o}dinger--F{\"o}llmer process
\citep{sfac2026}.

\subsection{Paper organization}
The remainder of the paper is organized as follows.
Section~\ref{sec:preliminaries} introduces the model, observations,
and four robust Bellman equations.
Section~\ref{sec:methodology} constructs the robust critic and conditional diffusion actor.
Section~\ref{sec:theory} gives their error and return guarantees.
 Section  \ref{sec:conclusion} concludes this work.
The detailed proofs are provided in the Appendix.

\section{Preliminaries}\label{sec:preliminaries}

\subsection{Notation}\label{sec:notations}
The state and action spaces are
$\mathcal X=[0,1]^{d_{\mathcal X}}$ and $\mathcal A=[0,1]^{d_{\mathcal A}}$, respectively,
where $d_{\mathcal X},d_{\mathcal A}$ are positive integers. Set
$d=d_{\mathcal X}+d_{\mathcal A}$. Write
$z=(x,a)$ for a state--action pair. The reward $r:\mathcal X\times
\mathcal A\to[0,\bar r]$ is known, with $\bar r>0$. The discount factor is
$\gamma\in(0,1)$, and $\nu$ denotes the initial state distribution. Set $B=\bar r/(1-\gamma)$,
the uniform bound on discounted values. For a function $f$,
$\|f\|_\infty$ denotes its supremum norm; $\|\cdot\|_2$ and
$\|\cdot\|_\infty$ on vectors denote Euclidean and maximum norms.
For a map $F$ and probability measure $\mu$, $F_\#\mu$ is the
distribution of $F(Y)$ when $Y\sim\mu$.
Write $\mathcal P(\mathcal X)$ for the set of probability measures on the state cube,
$\mathcal B(\mathcal X)$ for its Borel sigma-algebra, and
$C(\mathcal X)$ for the space of continuous real-valued functions on $\mathcal X$.
For a real number $u$, write $(u)_+=\max(u,0)$.
The notation $\widetilde{\mathcal O}$ suppresses multiplicative
polylogarithmic factors in the sample sizes.

For probability measures $P$ and $Q$ on $\mathcal X$, let
$\Gamma(P,Q)$ denote the set of their joint distributions with
marginals $P$ and $Q$. The $1$-Wasserstein distance is
\[
W_1(P,Q)=\inf_{\kappa\in\Gamma(P,Q)}
\int_{\mathcal X\times\mathcal X}\|x-y\|_2\,\kappa(dx,dy).
\]
The total variation distance is
\[
\operatorname{TV}(P,Q)=\|P-Q\|_{\mathrm{TV}}
=\sup_{E\in\mathcal B(\mathcal X)}|P(E)-Q(E)|.
\]
When $P$ and $Q$ have densities $p$ and $q$ with respect to a common
dominating measure $\mu$, this convention gives $\frac12\int|p-q|\,d\mu$.
If $Q\ll P$, meaning that $Q$ is absolutely continuous with respect
to $P$, the Kullback--Leibler and Pearson chi-square divergences are
\[
\operatorname{KL}(Q\Vert P)
=\int_{\mathcal X}\log\!\left(\frac{dQ}{dP}\right)\,dQ,
\qquad
\chi^2(Q\Vert P)
=\int_{\mathcal X}\left(\frac{dQ}{dP}-1\right)^2dP,
\]
where $dQ/dP$ is the Radon--Nikodym derivative. Both divergences
are defined as $+\infty$ when $Q\not\ll P$; the Pearson convention
has no factor $1/2$. The order $Q\Vert P$ specifies the direction
of these generally asymmetric divergences.
These conventions also apply on $\mathcal A$ and Euclidean spaces,
with $W_1$ restricted to probability measures with finite first moments.

\paragraph{ReLU DNNs.}
A ReLU deep neural network (DNN) with $J$ hidden layers has the form
\[
f_\theta(u)=A_J\circ\sigma\circ A_{J-1}\circ\cdots\circ\sigma\circ A_0(u),
\qquad A_j(v)=W_jv+b_j,
\]
where $\sigma(v)=\max(v,0)$ acts coordinatewise,
$W_j\in\mathbb R^{p_{j+1}\times p_j}$, $b_j\in\mathbb R^{p_{j+1}}$ for $0\leq j\leq J$,
and $\theta$ collects all weights and biases. The input and output dimensions
are $p_0$ and $p_{J+1}$. For prescribed dimensions, write
$\mathcal G(J,W,S,M)$ for these networks with hidden widths at most $W$,
at most $S$ nonzero parameters, and parameter magnitudes at most $M$.
Choose architecture budgets satisfying $J,W\leq S\leq JW^2$.

\paragraph{H\"older classes.}
For $\beta>0$, write $\beta=\ell+\vartheta$, where
$\ell=\lceil\beta\rceil-1$ and $\vartheta\in(0,1]$.
For a multi-index $\xi\in\mathbb N_0^p$, let
$|\xi|=\sum_{j=1}^p\xi_j$ and
$\partial^\xi=\partial_1^{\xi_1}\cdots\partial_p^{\xi_p}$.
For $\Sigma>0$, the class $H^\beta(\mathbb R^p,\Sigma)$ consists of
functions with continuous derivatives through order $\ell$ satisfying
\[
\max_{|\xi|\leq\ell}\|\partial^\xi f\|_\infty\leq\Sigma,
\qquad
\max_{|\xi|=\ell}\sup_{u\neq v}
\frac{|\partial^\xi f(u)-\partial^\xi f(v)|}
{\|u-v\|_2^\vartheta}\leq\Sigma.
\]
On a domain $U\subset\mathbb R^p$, $H^\beta(U,\Sigma)$ denotes
restrictions of these functions to $U$; $H^\beta(U)$ denotes the union
over finite $\Sigma$.

\subsection{Robust MDP}\label{sec:uncertainty}
A discounted MDP is specified by
$(\mathcal X,\mathcal A,P,r,\gamma,\nu)$, where
$P(\cdot\mid x,a)$ is the distribution of the next state conditional
on the current state $x$ and action $a$. A stationary stochastic
policy $\pi(\cdot\mid x)$ assigns a probability distribution on
$\mathcal A$ to each state.

\paragraph{Observations.}\label{sec:observations}
The nominal kernel $P_0$ is unknown. Before policy iteration,
observe one offline dataset
\[
\mathcal D_{\mathrm{off}}
=\bigl((X_i,A_i,R_i,X_i')\bigr)_{i=1}^{n_x},\qquad
(X_i,A_i)\sim\mu_{\mathrm{off}},\quad
R_i=r(X_i,A_i),\quad X_i'\sim P_0(\cdot\mid X_i,A_i).
\]
Here, $\mu_{\mathrm{off}}$ is the offline state--action design.
The model-estimation analysis assumes that these tuples are independent.
A conditional diffusion model fitted to this dataset defines
$\widehat P$, which is fixed throughout all actor--critic updates.

At round $k$, hold the implemented policy $\pi_k$ fixed and
collect $n$ fresh on-policy transitions,
\[
A_{k,i}\sim\pi_k(\cdot\mid X_{k,i}),\qquad
R_{k,i}=r(X_{k,i},A_{k,i}),\qquad
X'_{k,i}\sim P_0(\cdot\mid X_{k,i},A_{k,i}),\quad 1\leq i\leq n.
\]
Write $\mathcal D_k$ for the resulting dataset. It supplies the on-policy query design for estimating
$Q_D^{\pi_k}$. The successor observations in this batch are not
used to refit $\widehat P$ or to construct the model-based targets
below; transition estimation uses the original $n_x$ offline tuples.
For consecutive trajectory observations,
$X'_{k,i}=X_{k,i+1}$, so the tuples need not be independent.
For the analysis with independent observations, collect independent
nominal rollouts under $\pi_k$ from $\nu$ and retain one transition
at a common fixed time $t_{\mathrm{obs}}\geq1$ from each rollout.
Their state--action law is $\mu_k$, and the tuples are independent
conditional on the preceding fitted history. Using consecutive observations from each trajectory requires a separate
dependence analysis
\citep{feng2023,jiao2025}.
The reward is fixed and known; uncertainty concerns transitions.

\paragraph{Value functions.}
Write $\mathbb E_x^{\pi,P}$ for expectation when the trajectory starts
at $X_0=x$, and $\mathbb E_{x,a}^{\pi,P}$ when its first action is
also fixed at $A_0=a$ and subsequent actions follow $\pi$.
The state-value and action-value functions in environment $P$ are
\[
V^{\pi,P}(x)=
\mathbb E_x^{\pi,P}\left[\sum_{t=0}^{\infty}\gamma^tR_t\right],
\qquad
Q^{\pi,P}(x,a)=
\mathbb E_{x,a}^{\pi,P}\left[\sum_{t=0}^{\infty}\gamma^tR_t\right].
\]
For each choice $D\in\{W_1,\mathrm{TV},\mathrm{KL},\chi^2\}$, a prescribed
radius $\rho_D>0$ defines
$\mathcal U_D(x,a)=\mathcal U_D(P_0(\cdot\mid x,a);\rho_D)$.
Using the conventions in Section~\ref{sec:notations}, we define the
four sets and give their exact conditional duals.

\paragraph{Four uncertainty sets.}\label{sec:inner-solvers}

Fix a state--action pair and write \(P\) for its nominal next-state
distribution on \(\mathcal X=[0,1]^{d_{\mathcal X}}\).  Let
\(V:\mathcal X\to[0,B]\) be continuous, where \(B>0\).
Continuity suffices for minimization over the compact state space; no
bound on the Lipschitz constant of \(V\) is needed for these dual representations.
Here, \(P=P_0(\cdot\mid x,a)\) and \(\rho=\rho_D\) when used in Algorithm~\ref{form:algorithm}. For \(\rho\geq0\), define
\[
\mathcal U_{W_1}(P;\rho)=
 \left\{q:W_1(q,P)\leq\rho\right\},\qquad
\mathcal U_{\mathrm{TV}}(P;\rho)=
 \left\{q:\operatorname{TV}(q,P)\leq\rho\right\},
\]
\[
\mathcal U_{\mathrm{KL}}(P;\rho)=
 \left\{q:\operatorname{KL}(q\Vert P)\leq\rho\right\},\qquad
\mathcal U_{\chi^2}(P;\rho)=
 \left\{q:\chi^2(q\Vert P)\leq\rho\right\}.
\]
Here, all \(q\) are probability measures on \(\mathcal X\), with the
distances and divergences defined in Section~\ref{sec:notations}.
The TV and \(W_1\) sets permit mass outside the support of \(P\), whereas
the KL and \(\chi^2\) sets require \(q\ll P\).
For TV it suffices to consider \(\rho\leq1\), since every radius at least
one gives all probability measures on \(\mathcal X\).

The following formulas specialize standard distributional duality to
the present conventions; the \(W_1\) representation follows the transport duality of \citet{GaoKleywegt2023}.  The positive part in the Pearson formula is
essential: variance expressions require additional conditions
\citep{DuchiNamkoong2019}.  Robust RL under these four uncertainty families
has been studied in the tabular setting
\citep{XuPanagantiKalathil2023}.

\begin{lemma}[Exact conditional duals]\label{robdual:four}
For $P\in\mathcal P(\mathcal X)$, $v\in C(\mathcal X;[0,B])$,
and $\rho>0$, with $\rho\leq1$ for TV,

\begin{equation}
\inf_{q\in\mathcal U_{W_1}(P;\rho)}\int v\,dq
=\max_{0\leq\eta\leq B/\rho}
\left[\int\min_{y\in\mathcal X}
  \bigl(v(y)+\eta\|z-y\|_2\bigr)P(dz)-\eta\rho\right],
\label{form:dual-w1}
\end{equation}
\begin{equation}
\inf_{q\in\mathcal U_{\mathrm{TV}}(P;\rho)}\int v\,dq
=\max_{0\leq\tau\leq B}
\left[\int\min(v(z),\tau)P(dz)
-\rho\left(\tau-\min_{\mathcal X}v\right)\right],
\label{form:dual-tv}
\end{equation}
\begin{equation}
\inf_{q\in\mathcal U_{\mathrm{KL}}(P;\rho)}\int v\,dq
=\max_{0\leq\eta\leq B/\rho}
\left[-\eta\log\int e^{-v(z)/\eta}P(dz)-\eta\rho\right],
\label{form:dual-kl}
\end{equation}
\begin{equation}
\inf_{q\in\mathcal U_{\chi^2}(P;\rho)}\int v\,dq
=\max_{0\leq\tau\leq 2B(1+\rho)/\rho}
\left[\tau-\sqrt{(1+\rho)\int(\tau-v(z))_+^2P(dz)}\right].
\label{form:dual-chi}
\end{equation}
The KL expression at the endpoint $\eta=0$ equals $\essinf_P v$.
At radius zero all four values equal $\int v\,dP$.
\end{lemma}

\paragraph{Robust value functions and objective.}
The transition family is state--action rectangular:
\begin{equation*}
\mathfrak P_D=
\left\{P:\ P(\cdot\mid x,a)\in\mathcal U_D(x,a)
\text{ for every }(x,a)\right\}.
\end{equation*}
Here, $P$ ranges over transition kernels on $\mathcal X$. Thus the
adversary may choose different admissible conditional laws at different
state--action pairs. The radius is a model parameter, rather than a
claim that a ball around an empirical distribution covers $P_0$.
Different distances are studied separately; equal radius values need
not describe comparable perturbations. Each uncertainty family uses a
fixed positive radius; the TV radius is at most one.

For each uncertainty type $D$, define the robust state and action values
\begin{equation}
V_D^\pi(x)=\inf_{P\in\mathfrak P_D}V^{\pi,P}(x),
\qquad
Q_D^\pi(x,a)=\inf_{P\in\mathfrak P_D}Q^{\pi,P}(x,a).
\label{eq:robust-values-definition}
\end{equation}
The infimum in each definition
uses one admissible stationary kernel throughout the trajectory.
For the initial law $\nu$, the robust return is
\begin{equation}
J_D(\pi)=
\inf_{P\in\mathfrak P_D}
\mathbb E_\nu^{\pi,P}
\left[\sum_{t=0}^{\infty}\gamma^tR_t\right].
\label{form:objective}
\end{equation}
The learning objective is $\sup_\pi J_D(\pi)$ for the chosen $D$.
The optimal robust values are
$V_D^\star(x)=\sup_\pi V_D^\pi(x)$ and
$Q_D^\star(x,a)=\sup_\pi Q_D^\pi(x,a)$.

\paragraph{Robust Bellman equations.}\label{sec:bellman}
Under Assumption~\ref{ass:cdm-regularity}, the weakly continuous
policies used by the algorithm satisfy
\begin{equation}
V_D^\pi(x)=\int_{\mathcal A}Q_D^\pi(x,a)\pi(da\mid x),\qquad
Q_D^\pi(x,a)=r(x,a)+\gamma\inf_{q\in\mathcal U_D(x,a)}\int_{\mathcal X}V_D^\pi(y)q(dy).
\label{eq:bellman-policy}
\end{equation}
The optimal Bellman equations are
\[
V_D^\star(x)=\sup_{a\in\mathcal A}Q_D^\star(x,a),\qquad
Q_D^\star(x,a)=r(x,a)+\gamma\inf_{q\in\mathcal U_D(x,a)}\int_{\mathcal X}\sup_{b\in\mathcal A}Q_D^\star(y,b)\,q(dy).
\]
For continuous $v$, let $\mathcal T_D^\pi v$ be the right side of
the state-value equation obtained from \eqref{eq:bellman-policy}
by replacing $V_D^\pi$ with $v$. Let $\mathcal T_Dv$ replace the
policy integral by the supremum over actions. Thus
$V_D^\pi=\mathcal T_D^\pi V_D^\pi$ and
$V_D^\star=\mathcal T_DV_D^\star$.
Substitution of Lemma~\ref{robdual:four} gives the following
four pairs of explicit equations. In each pair, the bracketed quantity
is $Q_D^\pi(x,a)$ or $Q_D^\star(x,a)$, respectively.

\paragraph{Wasserstein uncertainty.}
For $D=W_1$, the policy-evaluation and optimality equations are
\begin{align*}
&V_{W_1}^\pi(x)\\
&=\int_{\mathcal A}\Biggl[r(x,a)+\gamma
\max_{0\le\lambda\le B/\rho_{W_1}}
\biggl(\int_{\mathcal X}\min_{z\in\mathcal X}
\bigl[V_{W_1}^\pi(z)+\lambda\|y-z\|_2\bigr]P_0(dy\mid x,a)-\lambda\rho_{W_1}\biggr)
\Biggr]\pi(da\mid x),\\
&V_{W_1}^\star(x)\\
&=\sup_{a\in\mathcal A}\Biggl[r(x,a)+\gamma
\max_{0\le\lambda\le B/\rho_{W_1}}
\biggl(\int_{\mathcal X}\min_{z\in\mathcal X}
\bigl[V_{W_1}^\star(z)+\lambda\|y-z\|_2\bigr] P_0(dy\mid x,a)-\lambda\rho_{W_1}\biggr)
\Biggr].
\end{align*}
The transport minimization ranges over the entire next-state cube.

\paragraph{Total-variation uncertainty.}
For $D=\mathrm{TV}$ and $0<\rho_{\mathrm{TV}}\le1$,
\begin{align*}
V_{\mathrm{TV}}^\pi(x)
&=\int_{\mathcal A}\Biggl[r(x,a)+\gamma\max_{0\le t\le B}
\biggl(\int_{\mathcal X}\min\bigl[V_{\mathrm{TV}}^\pi(y),t\bigr]
\\
&\hspace{20mm}P_0(dy\mid x,a)-\rho_{\mathrm{TV}}\bigl[t-\min_{z\in\mathcal X}V_{\mathrm{TV}}^\pi(z)\bigr]
\biggr)\Biggr]\pi(da\mid x),\\
V_{\mathrm{TV}}^\star(x)
&=\sup_{a\in\mathcal A}\Biggl[r(x,a)+\gamma\max_{0\le t\le B}
\biggl(\int_{\mathcal X}\min\bigl[V_{\mathrm{TV}}^\star(y),t\bigr]
 P_0(dy\mid x,a)-\rho_{\mathrm{TV}}\bigl[t-\min_{z\in\mathcal X}V_{\mathrm{TV}}^\star(z)\bigr]
\biggr)\Biggr].
\end{align*}
The global state minimum in \eqref{form:dual-tv} reflects the ability to relocate admissible
mass to a least favorable state.

\paragraph{Kullback--Leibler uncertainty.}
For $D=\mathrm{KL}$,
\begin{align*}
V_{\mathrm{KL}}^\pi(x)
&=\int_{\mathcal A}\Biggl[r(x,a)+\gamma
\max_{0\le\eta\le B/\rho_{\mathrm{KL}}}
\biggl(-\eta\log\int_{\mathcal X}e^{-V_{\mathrm{KL}}^\pi(y)/\eta}
P_0(dy\mid x,a)-\eta\rho_{\mathrm{KL}}\biggr)\Biggr]\pi(da\mid x),\\
V_{\mathrm{KL}}^\star(x)
&=\sup_{a\in\mathcal A}\Biggl[r(x,a)+\gamma
\max_{0\le\eta\le B/\rho_{\mathrm{KL}}}
\biggl(-\eta\log\int_{\mathcal X}e^{-V_{\mathrm{KL}}^\star(y)/\eta}
P_0(dy\mid x,a)-\eta\rho_{\mathrm{KL}}\biggr)\Biggr].
\end{align*}
In \eqref{form:dual-kl}, the penalized logarithmic expression at the
endpoint $\eta=0$ is interpreted as the essential infimum of the relevant
value function under
$P_0(\cdot\mid x,a)$. Under Assumption~\ref{ass:cdm-regularity},
this equals its minimum over $\mathcal X$.

\paragraph{Pearson chi-square uncertainty.}
For $D=\chi^2$,
\begin{align*}
&V_{\chi^2}^\pi(x)\\
&=\int_{\mathcal A}\Biggl[r(x,a)+\gamma
\max_{0\le t\le 2B(1+\rho_{\chi^2})/\rho_{\chi^2}}
\biggl(t-\sqrt{(1+\rho_{\chi^2})
\int_{\mathcal X}(t-V_{\chi^2}^\pi(y))_+^2
P_0(dy\mid x,a)}\biggr)\Biggr]\pi(da\mid x),\\
&V_{\chi^2}^\star(x)\\
&=\sup_{a\in\mathcal A}\Biggl[r(x,a)+\gamma
\max_{0\le t\le 2B(1+\rho_{\chi^2})/\rho_{\chi^2}}
\biggl(t-\sqrt{(1+\rho_{\chi^2})
\int_{\mathcal X}(t-V_{\chi^2}^\star(y))_+^2
P_0(dy\mid x,a)}\biggr)\Biggr].
\end{align*}
The positive part enforces the nonnegativity
of the alternative transition density.

Each policy-evaluation operator $\mathcal T_D^\pi$ and optimality
operator $\mathcal T_D$ is a $\gamma$-contraction on the relevant
continuous-function space. Lemma~\ref{prop:rsfac-bellman}
identifies their fixed points with \eqref{eq:robust-values-definition}
and gives $J_D(\pi)=\int V_D^\pi\,d\nu$ and
$\sup_\pi J_D(\pi)=\int V_D^\star\,d\nu$.
An optimal stationary policy can be selected from
$\operatorname*{arg\,max}_{a\in\mathcal A}Q_D^\star(x,a)$.
The resulting optimal value also bounds history-dependent policies.

\subsection{Conditional diffusion models}\label{sec:cdm-preliminaries}
We describe conditional diffusion models through forward and backward
stochastic differential equations (SDEs).
Let $Y_0$ take values in $\mathcal Y=[0,1]^{d_{\mathcal Y}}$ and have
conditional density $p_0(\cdot\mid z)$ given
$Z=z\in[0,1]^{d_{\mathcal Z}}$.
The reference scores are ReLU DNNs in $\mathcal G(J,W,S,M)$ with
input $(t,y,z)\in\mathbb R^{1+d_{\mathcal Y}+d_{\mathcal Z}}$
and output in $\mathbb R^{d_{\mathcal Y}}$.
Their architecture and output bounds are specified in Section~\ref{sec:theory}.

\paragraph{Forward SDE.}
Let $B_t$ be a standard $d_{\mathcal Y}$-dimensional Brownian motion
independent of $(Y_0,Z)$. The forward Ornstein--Uhlenbeck (OU) SDE is
\begin{equation}
dY_t=-\tfrac12Y_t\,dt+dB_t,\qquad Y_0\mid Z=z\sim p_0(\cdot\mid z),\quad 0\leq t\leq T.
\label{eq:cdm-forward}
\end{equation}
For each $t>0$, its conditional marginal distribution has the representation
\begin{equation}
Y_t\overset{d}=e^{-t/2}Y_0+\sqrt{1-e^{-t}}\,G,\qquad G\sim N(0,I_{d_{\mathcal Y}}),
\label{eq:cdm-marginal}
\end{equation}
where $G$ is independent of $(Y_0,Z)$. Thus
$Y_t\mid Z=z\Rightarrow N(0,I_{d_{\mathcal Y}})$ as $t\to\infty$.
Write $p_t(\cdot\mid z)$ for this density. Its conditional score is
\begin{equation}
s_t(y\mid z)=\nabla_y\log p_t(y\mid z)
=\frac{e^{-t/2}\mathbb E[Y_0\mid Y_t=y,Z=z]-y}{1-e^{-t}}.
\label{eq:cdm-score}
\end{equation}
For $0<\tau<T/2$, denoising score matching estimates this score by
minimizing an empirical version of
\begin{equation}
\frac1{T-2\tau}\int_\tau^{T-\tau}\mathbb E
\left\|s\!\left(t,e^{-t/2}Y_0+\sqrt{1-e^{-t}}G,Z\right)
+\frac{G}{\sqrt{1-e^{-t}}}\right\|_2^2\,dt.
\label{eq:cdm-denoising}
\end{equation}

\paragraph{Backward SDE.}
Conditional on $Z=z$, the time-reversed process $\bar Y_u=Y_{T-u}$
starts from $p_T(\cdot\mid z)$ and satisfies
\begin{equation}
d\bar Y_u=\bigl[\tfrac12\bar Y_u+s_{T-u}(\bar Y_u\mid z)\bigr]du+d\bar B_u,
\qquad 0\leq u<T,
\label{eq:cdm-backward}
\end{equation}
where $\bar B_u$ is Brownian motion for the reversed filtration.
Its conditional marginal is $p_{T-u}(\cdot\mid z)$.
For generation, replace the score by its estimate $\widehat s$, the
initial law at forward time $T-\tau$ by $N(0,I_{d_{\mathcal Y}})$,
and use the interval $[\tau,T-\tau]$. With $L\geq1$ steps and
$\Delta=(T-2\tau)/L$, the finite sampler is
\begin{equation}
\widehat Y_{\ell+1}=\widehat Y_\ell
+\Delta\bigl[\tfrac12\widehat Y_\ell+\widehat s(T-\tau-\ell\Delta,\widehat Y_\ell,z)\bigr]
+\sqrt{\Delta}\,\xi_\ell,\qquad 0\leq\ell<L.
\label{eq:cdm-em}
\end{equation}
Here, $\widehat Y_0,\xi_0,\ldots,\xi_{L-1}$ are independent standard
Gaussian vectors. After the $L$ steps in \eqref{eq:cdm-em}, the raw output
$\widehat Y_L$ may lie outside $\mathcal Y$. The final output is
$\mathsf F(\widehat Y_L)\in\mathcal Y$, where the coordinatewise folding map is
\begin{equation}
\mathsf F(y)_j=\operatorname{dist}(y_j,2\mathbb Z).
\label{form:cube-retraction}
\end{equation}
This map is $1$-Lipschitz and is the identity on the cube; $\mathsf F_q$ denotes
its $q$-dimensional version.
Section~\ref{form:critic-section} uses the next state as the response and
the current state--action pair $Z=(X,A)$ as the conditioning variable, with
$(d_{\mathcal Y},d_{\mathcal Z})=(d_{\mathcal X},d)$.
Section~\ref{form:actor-section} applies the same construction to
weighted, augmented state--action samples for policy improvement, with
$(d_{\mathcal Y},d_{\mathcal Z})=(d_{\mathcal A},d_{\mathcal X})$.

\section{Methodology}\label{sec:methodology}\label{form:algorithm-section}
In this section, we develop an algorithm that alternates robust
action-value estimation with conditional diffusion policy improvement.

Fix one uncertainty type $D$ and write $\rho=\rho_D$ throughout the iteration;
the four choices share this construction but generally produce
different policies.
Write $Q_k=Q_D^{\pi_k}$ for the true robust action-value function
under the current policy and $\widehat Q_k$ for its estimator. The current
policy's robust Bellman operator $\mathcal B_k$ is defined in \eqref{form:true-q-operator}.
Fix $0<\epsilon<1$ and $0<\lambda\leq1$ within one run.
The ideal relative policy update maximizes
\[
\int Q_k(x,a)\pi(da\mid x)
-\lambda\KL(\pi(\cdot\mid x)\Vert\pi_k(\cdot\mid x))
\]
at each state, exponentially tilting $\pi_k$ by $Q_k$.
The implemented actor further regularizes this target as described
in Section~\ref{form:actor-section}.

Starting from an initial policy $\pi_0$, at round $k$ the critic estimates
$Q_k$ for the actual current policy, yielding $\widehat Q_k$.
A nominal model $\widehat P$ is learned once from
$\mathcal D_{\mathrm{off}}$ before policy iteration and remains fixed.
The estimated relative target is
\begin{equation}
\bar\pi_{k+1}(da\mid x)
=\frac{\exp(\widehat Q_k(x,a)/\lambda)\pi_k(da\mid x)}
{\int_{\mathcal A}\exp(\widehat Q_k(x,b)/\lambda)\pi_k(db\mid x)}.
\label{form:relative-target}
\end{equation}
Subtracting a state-only baseline leaves this distribution unchanged,
so the actor uses $\widehat Q_k$ directly. Weighted conditional
score estimation, finite reverse diffusion, and the stated policy
regularization define the next actual policy $\pi_{k+1}$.
Thus the two components are linked by
\[
(\pi_k,\mathcal D_k;\widehat P)
\xrightarrow{\text{robust critic}}\widehat Q_k
\xrightarrow{\eqref{form:relative-target}}\bar\pi_{k+1}
\xrightarrow{\text{weighted CDM actor}}\pi_{k+1}.
\]
Both the next critic and the next relative update use $\pi_{k+1}$.
The temperature regularizes policy improvement; the critic continues
to evaluate the unregularized robust return in \eqref{form:objective}.

Algorithm~\ref{form:algorithm} summarizes the robust actor--critic procedure.
The final critic fit evaluates the returned policy. This convention uses
$K$ actor updates and $K+1$ critic fits.

\begin{algorithm}[H]
\caption{Robust actor--critic with conditional diffusion models}
\label{form:algorithm}
\begin{algorithmic}[1]
\REQUIRE Initial policy $\pi_0$, offline data $\mathcal D_{\mathrm{off}}$ of size $n_x$, per-round critic sample size $n$, common sample size $m$, reward $r$,
uncertainty type $D$ and radius $\rho$, discount $\gamma$,
temperature $\lambda$, exploration probability $\epsilon$, actor network
resolution $N$ and $K$ policy updates.
\STATE Fit $\widehat P$ once from $\mathcal D_{\mathrm{off}}$ by
\eqref{eq:cdm-denoising} and~\eqref{form:learned-kernel} and keep it fixed.
\FOR{$k=0,\ldots,K$}
  \STATE Freeze the actual current-policy sampler $\pi_k$.
  \STATE Collect $n$ fresh critic tuples $\mathcal D_k$ under
  $\pi_k$ as in Section~\ref{sec:observations}.
  \STATE Construct robust targets by \eqref{form:robust-label}
  and fit $\widehat Q_k$ over $\mathcal Q$ using
  \eqref{form:critic-fit} with auxiliary DNN class $\mathcal O$.
  \IF{$k<K$}
    \STATE Apply Algorithm~\ref{form:actor-algorithm} to
    $\pi_k$ and $\widehat Q_k$ to define $\pi_{k+1}$.
  \ENDIF
\ENDFOR
\ENSURE The implemented policy $\pi_K$ and its robust critic
$\widehat Q_K$.
\end{algorithmic}
\end{algorithm}

\subsection{Robust critic estimation}\label{form:critic-section}
Use a common sample size $m$ for critic integration and actor fitting:
each query has $m$ generated
successors, the critic action integral uses an independent bank
of $m$ draws, and each actor update uses a separate training sample of $m$
current-policy queries. These samples are independent.

\paragraph{Nominal transition learning.}\label{form:transition-model}
Apply Section~\ref{sec:cdm-preliminaries} to $\mathcal D_{\mathrm{off}}$,
with response $Y_0=X'$ and conditioning variable $Z=(X,A)$.
Using the forward perturbation \eqref{eq:cdm-marginal}, fit
$\widehat s$ by the empirical denoising criterion
\eqref{eq:cdm-denoising} based on the $n_x$ offline tuples and the
independent auxiliary perturbations specified in Section~\ref{sec:theory}.
At a query $(x,a)$, use \eqref{eq:cdm-em} with the state as the response,
conditioning value $z=(x,a)$, $T=T_0$, and $\tau=t_{\min}$,
where $0<t_{\min}<T_0/2$.
Let $G_{\widehat s}(x,a,\zeta)$ be the finite sampler, including
the final map $\mathsf F$ on the state space, where $\zeta$ denotes fresh
sampling randomness. Writing $\mathcal L$ for the conditional law,
define the implemented transition kernel by
\begin{equation}
\widehat P(\cdot\mid x,a)
=\mathcal L\bigl(G_{\widehat s}(x,a,\zeta)
\mid\mathcal D_{\mathrm{off}},\widehat s\bigr).
\label{form:learned-kernel}
\end{equation}
 
\paragraph{Robust Bellman targets.}
For a continuous candidate $Q:\mathcal X\times\mathcal A\to[0,B]$,
the true policy-evaluation operator is
\begin{equation}
(\mathcal B_kQ)(x,a)
=r(x,a)+\gamma
\inf_{q\in\mathcal U_D(P_0(\cdot\mid x,a);\rho)}
\int_{\mathcal X}\int_{\mathcal A}Q(y,b)\pi_k(db\mid y)q(dy).
\label{form:true-q-operator}
\end{equation}
Its fixed point is $Q_k$.
Write $\widehat{\mathcal B}_k$ for the operator obtained
from \eqref{form:true-q-operator} by replacing $P_0$ with
\eqref{form:learned-kernel}.

Use the fresh on-policy query batch $\mathcal D_k$ of size $n$
with design $\mu_k$. At each round, draw a fresh bank of independent,
identically distributed noises for action generation, independently of
the preceding fitted history, fresh query batch, and successor-generation
noises.
Let $\mathcal H_k$ contain the offline data,
fixed fitted model, preceding policy history, and this bank. Conditional on $\mathcal H_k$, the fresh query
pairs $(x_{k,i},a_{k,i})$, $i=1,\ldots,n$, are independent with
law $\mu_k$. For each query, generate
$m$ independent successors $\widetilde X'_{k,i,\ell}$ from
$\widehat P(\cdot\mid x_{k,i},a_{k,i})$. These batches are independent
across $i$ conditional on $\mathcal H_k$ and all query pairs. Form the empirical
measure
\[
\widetilde P_{ki}
=\frac1m\sum_{\ell=1}^m\delta_{\widetilde X'_{k,i,\ell}}.
\]
For each candidate $Q$, first average over actions under the fixed
current policy. Let $A_k(y,\zeta)$ denote its implemented action
sampler, and use a common independent noise bank to define
\[
\widetilde V_{k,Q}(y)=\frac1{m}\sum_{j=1}^{m}
Q\bigl(y,A_k(y,\zeta_{k,j})\bigr),\qquad
A_k(y,\zeta_{k,j})\sim\pi_k(\cdot\mid y).
\]
The bank provides $m$ action draws per state, is independent of the
successor samples, and remains fixed across candidate critics.

For $\rho>0$, apply the four conditional duals to
$\widetilde P_{ki}$. The scalar parameter intervals are
\[
\mathcal I_{W_1}=[0,B/\rho],\qquad
\mathcal I_{\mathrm{TV}}=[0,B],\qquad
\mathcal I_{\mathrm{KL}}=[0,B/\rho],\qquad
\mathcal I_{\chi^2}=[0,2B(1+\rho)/\rho].
\]
For $u$ in the corresponding interval, define the empirical dual
objective functions
\[
F_{W_1,i}(u;Q)=\frac1m\sum_{\ell=1}^m
\min_{y\in\mathcal X}\left[
\widetilde V_{k,Q}(y)+u\|\widetilde X'_{k,i,\ell}-y\|_2\right]
-u\rho,
\]
\[
F_{\mathrm{TV},i}(u;Q)=\frac1m\sum_{\ell=1}^m
\min\left[\widetilde V_{k,Q}(\widetilde X'_{k,i,\ell}),u\right]
-\rho\left[u-\min_{y\in\mathcal X}\widetilde V_{k,Q}(y)\right],
\]
\[
F_{\mathrm{KL},i}(u;Q)
=-u\log\!\left[\frac1m\sum_{\ell=1}^m
\exp\!\left(-\frac{\widetilde V_{k,Q}(\widetilde X'_{k,i,\ell})}{u}\right)\right]
-u\rho,\qquad u>0,
\]
\[
F_{\chi^2,i}(u;Q)=u-
\sqrt{\frac{1+\rho}m\sum_{\ell=1}^m
\left[u-\widetilde V_{k,Q}(\widetilde X'_{k,i,\ell})\right]_+^2}.
\]
The KL endpoint is
$F_{\mathrm{KL},i}(0;Q)
=\min_{1\leq\ell\leq m}\widetilde V_{k,Q}(\widetilde X'_{k,i,\ell})$.
The exact empirical-center target is therefore
\begin{equation}
Y_{k,i}(Q)
=r(x_{k,i},a_{k,i})+\gamma\max_{u\in\mathcal I_D}F_{D,i}(u;Q).
\label{form:robust-label}
\end{equation}
At $\rho=0$, set
\[
Y_{k,i}(Q)=r(x_{k,i},a_{k,i})
+\frac\gamma m\sum_{\ell=1}^m
\widetilde V_{k,Q}(\widetilde X'_{k,i,\ell}),
\]
so no interval involving $1/\rho$ is needed. For TV and $W_1$, the displayed state minima
still range over the full cube, including points outside the
generated sample support. The empirical
KL endpoint is defined on the empirical support.

The target takes values in $[0,B]$ and depends on
the candidate critic $Q$ through $\widetilde V_{k,Q}$ and the dual
expression. The empirical center approximates the learned law $\widehat P$,
whereas the statistical target remains the robust problem centered
at the true $P_0$. Finite successor sampling and
action averaging contribute to the conditional-mean target bias.

\paragraph{Minimax Bellman estimation.}
Use the $[0,B]$-valued ReLU DNNs in $\mathcal G(J,W,S,M)$ for both
$\mathcal Q$ and $\mathcal O$, with input dimension $d$ and scalar output.
Their uniform Lipschitz bound and architectures are specified in
Lemma~\ref{thm:final-critic-rate}.
For $Q\in\mathcal Q$ and $O\in\mathcal O$, the population
minimax criterion is
\begin{equation}
\min_{Q\in\mathcal Q}\max_{O\in\mathcal O}
\int\left[
\bigl(Q-\mathcal B_kQ\bigr)^2
-\bigl(O-\mathcal B_kQ\bigr)^2
\right]\,d\mu_k.
\label{form:critic-population}
\end{equation}
For a fixed $Q$, the inner value in \eqref{form:critic-population} is
\[
\|Q-\mathcal B_kQ\|_{L^2(\mu_k)}^2
-\inf_{O\in\mathcal O}
\|O-\mathcal B_kQ\|_{L^2(\mu_k)}^2.
\]
Replacing $\mathcal B_k$ by
$\widehat{\mathcal B}_k$ in
\eqref{form:critic-population} defines the intermediate
learned-model population criterion. The difference between the two
criteria retains the transition-center estimation error.

Using the same target $Y_{k,i}(Q)$ in both squares, define the
critic through the empirical minimax criterion
\begin{equation}
\min_{Q\in\mathcal Q}\max_{O\in\mathcal O}
\frac1{n}\sum_{i=1}^{n}
\left[
\bigl(Q(x_{k,i},a_{k,i})-Y_{k,i}(Q)\bigr)^2
-\bigl(O(x_{k,i},a_{k,i})-Y_{k,i}(Q)\bigr)^2
\right].
\label{form:critic-fit}
\end{equation}
Let $\widehat Q_k$ be a minimizer of this criterion.
For fixed $Q$, the conditional variance cancels when both terms
are centered at
$\mathbb E[Y_{k,i}(Q)\mid x_{k,i},a_{k,i},\mathcal H_k]$.
The auxiliary regression therefore fits this conditional mean. Recovering the true Bellman residual additionally requires
control of its discrepancy from
$(\widehat{\mathcal B}_kQ)(x_{k,i},a_{k,i})$
and of the discrepancy between $\widehat P$ and $P_0$.
This is a current-policy Bellman-residual fit: $\pi_k$ and
$\widehat P$ remain fixed while both occurrences of
$Y_{k,i}(Q)$ are recomputed from the current candidate $Q$.

\subsection{Conditional diffusion policy improvement}\label{form:actor-section}
At each update, the actor learns a smoothed version of the relative
conditional target from current-policy queries. Gaussian augmentation
ensures that the target satisfies the conditional-diffusion regularity
conditions even with a ReLU critic. Gaussian postprocessing and
uniform exploration provide the policy regularity needed for
subsequent critic fits.

Choose $0<\epsilon<1$ and $0<\lambda\leq1$, keep them fixed
throughout the $K$ policy updates, and set
$\eta=\epsilon(1+B/\lambda)^{-1/2}$.
Choose the actor network resolution $N$ before the sample size $m$.
The allocations used for the error guarantees are specified in
Section~\ref{sec:theory}.

\paragraph{Weighted training samples.}
Condition on the fitted history and $\widehat Q_k$. Draw fresh
$X_j\sim\Unif(\mathcal X)$ and
$A_j\sim\pi_k(\cdot\mid X_j)$, $1\leq j\leq m$, and compute
\[
w_j=\exp\!\left(\widehat Q_k(X_j,A_j)/\lambda\right).
\]
Using independent standard Gaussians, augment these pairs by
\[
X_j^\eta=\mathsf F_{d_{\mathcal X}}(X_j+\eta G_{x,j}),\qquad
A_j^\eta=\mathsf F_{d_{\mathcal A}}(A_j+\eta G_{a,j}).
\]

\paragraph{Conditional score learning.}
Use \eqref{eq:cdm-forward}--\eqref{eq:cdm-marginal} with the augmented
action as the response and the augmented state as the conditioning variable.
The weighted empirical form of \eqref{eq:cdm-denoising} is
\begin{equation}
\frac1{m(T-2\tau)}\sum_{j=1}^m w_j
\int_\tau^{T-\tau}\mathbb E_G\left\|
s\!\left(t,e^{-t/2}A_j^\eta+\sqrt{1-e^{-t}}G,X_j^\eta\right)
+\frac{G}{\sqrt{1-e^{-t}}}\right\|_2^2\,dt.
\label{form:actor-score-fit}
\end{equation}
Use the network class, independent auxiliary perturbation budget,
time truncation parameters, and sampling resolution specified in
Section~\ref{sec:theory}, with response dimension
$d_{\mathcal A}$, conditioning dimension $d_{\mathcal X}$, and network resolution $N$.
The weighted, augmented joint law defines the conditional action
density $q_{k,\eta}(\cdot\mid x)$, the actor's smoothed target. Weighting the entire squared loss makes its unrestricted
population minimizer the conditional score of this target.

Using the score identity \eqref{eq:cdm-score}, apply to every
reference score the fixed support projection
\[
s(t,y,x)\longmapsto
\frac{e^{-t/2}\Pi_{[0,1]^{d_{\mathcal A}}}
\left(e^{t/2}\bigl[(1-e^{-t})s(t,y,x)+y\bigr]\right)-y}
{1-e^{-t}}.
\]
Denote the minimizer over this projected class by $\widehat s_{k+1}$.
Use a deterministic reverse grid with maximum step $\Delta\leq1$.
Section~\ref{sec:theory} gives the mesh required for the error guarantees.

\paragraph{Finite conditional generation.}
For a context $u$, use the backward dynamics
\eqref{eq:cdm-backward} and finite sampler \eqref{eq:cdm-em}, with
score $\widehat s_{k+1}$, conditioning value $z=u$, and response
dimension $d_{\mathcal A}$. Fold the output by
\eqref{form:cube-retraction} and denote its conditional law by
$\widehat\pi_{k+1}(\cdot\mid u)$.
At a query $x$, independently draw a Bernoulli variable with
success probability $\epsilon$. On success, return a uniform
action. Otherwise, draw independent standard Gaussians and set
\[
U=\mathsf F_{d_{\mathcal X}}(x+\epsilon G_x),\qquad
A^{\rm raw}\sim\widehat\pi_{k+1}(\cdot\mid U),\qquad
A=\mathsf F_{d_{\mathcal A}}(A^{\rm raw}+\epsilon G_a).
\]
The law of $A$ defines the implemented policy $\pi_{k+1}$ used in
the next critic fit and weighted update.

Algorithm~\ref{form:actor-algorithm} summarizes the conditional diffusion
policy-improvement step.

\begin{algorithm}[H]
\caption{Weighted conditional diffusion policy improvement}
\label{form:actor-algorithm}
\begin{algorithmic}[1]
\REQUIRE Current-policy sampler $\pi_k$, fitted critic $\widehat Q_k$,
sample size $m$, network resolution $N$, parameters $\epsilon,\lambda$,
time truncation parameters $\tau,T$, and number of reverse steps $L$.
\STATE Set $\eta=\epsilon(1+B/\lambda)^{-1/2}$ and $\Delta=(T-2\tau)/L$.
\STATE Draw fresh current-policy state--action pairs, evaluate their original
weights, and form the augmented contexts and actions.
\STATE Fit the support-consistent score by \eqref{form:actor-score-fit}
with the independent perturbation average prescribed by the CDM result.
\STATE At each query, augment the context, generate the raw action by
\eqref{eq:cdm-em}, and apply action augmentation and uniform exploration.
\ENSURE The  conditional sampler $\pi_{k+1}$.
\end{algorithmic}
\end{algorithm}

\section{Theoretical guarantees}\label{sec:theory}
In this section, we establish finite-sample guarantees for the proposed robust
actor--critic method. We state the density and coverage conditions and
specify the score classes and sampling allocations, then bound the
critic's on-policy $L^2$ error and the actor's conditional distribution
error. Combining these bounds with robust policy improvement yields
last-iterate return guarantees that retain nominal-model estimation,
target integration, and policy regularization errors. The analysis uses
the following regularity and coverage conditions.

\begin{assumption}[Transition regularity]\label{ass:cdm-regularity}
The reward is $L_r$-Lipschitz in the maximum norm.
For some $\alpha,\Sigma,\Sigma_0,p_-,p_+>0$ and $a_0\in(0,1/2)$,
the nominal density has the following properties:
\begin{enumerate}[label=(\roman*),leftmargin=*]
\item For every $z=(x,a)\in[0,1]^d$, $p_0(\cdot\mid z)$ is supported
on $\mathcal X$ and $p_-\leq p_0(y\mid z)\leq p_+$ there.
\item The joint function $(y,z)\mapsto p_0(y\mid z)$ belongs to
$H^\alpha([0,1]^{d_{\mathcal X}+d},\Sigma)$.
\item Its restriction to
$([0,1]^{d_{\mathcal X}}\setminus[a_0,1-a_0]^{d_{\mathcal X}})\times[0,1]^d$
admits a Whitney extension in
$H^{3\alpha+2}([-1,1]^{d_{\mathcal X}}\times[0,1]^d,\Sigma_0)$.
\item For every $\xi\in\mathbb N_0^d$ with
$|\xi|\leq\lfloor6\alpha+3\rfloor+1$, the derivative
$\partial_z^\xi p_0(y\mid z)$ exists and
$\sup_{(y,z)\in[0,1]^{d_{\mathcal X}+d}}|\partial_z^\xi p_0(y\mid z)|<\infty$.
\end{enumerate}
\end{assumption}

\begin{assumption}[Offline design coverage]\label{ass:fixed-model-coverage}
For the finite budgets under consideration, there is a deterministic
constant $C\geq1$ such that almost surely
\[
\mu_k(E)\leq C\mu_{\mathrm{off}}(E)
\quad\text{for every Borel }E\subseteq\mathcal X\times\mathcal A
\text{ and every round under consideration}.
\]
\end{assumption}

\begin{remark}
The nominal density conditions (i)--(iv) in
Assumption~\ref{ass:cdm-regularity} are Assumptions~4.1--4.4 of
\citet{Chen2026}, expressed in the present notation.
The bounded first conditioning derivatives in (iv) imply
\[
|p_0(y\mid z)-p_0(y\mid z')|
\leq\left(\sum_{j=1}^d\sup_{y,u}|\partial_{u_j}p_0(y\mid u)|\right)
\|z-z'\|_\infty
\leq L_p\|z-z'\|_\infty
\]
for a finite $L_p$.

Assumption~\ref{ass:fixed-model-coverage} compares the on-policy query
designs with the offline design; it does not constrain adversarial
transition laws or follow from continuity of the fitted diffusion alone.
Under the density bound in Assumption~\ref{ass:cdm-regularity},
Lemma~\ref{app:cdm-density} gives an explicit sufficient coverage constant
when the offline query density is bounded away from zero. 
\end{remark}

\paragraph{Sampler regularity.}
The initial policy has density $\pi_0(a\mid x)$ uniformly bounded above and away
from zero, with bounds independent of sample sizes. Choose $\epsilon$
below its density lower bound. Its common-noise sampler satisfies the
state Lipschitz moment bound below.
For the implemented sampler $A_k$, conditional on the fitted history
before round $k$, the squared state Lipschitz constant of
$A_k(\cdot,\zeta)$ has expectation at most $L_\pi^2$, almost surely,
for a deterministic $L_\pi$. States use the maximum norm and actions the
Euclidean norm; fresh sampling noise is independent of the critic data.
For fixed architecture, parameter bounds, positive cutoff, and finite
mesh, such a bound follows from Lemma~\ref{app:cdm-density}.
Consequently, the shared action bank in Section~\ref{form:critic-section}
defines a continuous $[0,B]$-valued function on the next-state cube.
The nominal model is TV-continuous in $(x,a)$: its final positive-variance
Gaussian step has TV-continuous translated densities, and the fixed
folding map preserves this property.

\paragraph{Score classes and sampling.}
For response dimension $d_{\mathcal Y}$, conditioning dimension
$d_{\mathcal Z}$, and sufficiently large resolution $N$, use the reference
score class $\mathcal G(J,W,S,M)$ with
\begin{align*}
J&\lesssim(\log N)^4,
&W&=\widetilde{\mathcal O}(N^{d_{\mathcal Y}+d_{\mathcal Z}}),\\
S&\lesssim N^{d_{\mathcal Y}+d_{\mathcal Z}}(\log N)^9,
&\log M&\lesssim(\log N)^4.
\end{align*}
Response inputs are clipped at a sufficiently large multiple of
$\sqrt{\log N}$, and the outputs satisfy
\[
\sup_{y,z}\|s(t,y,z)\|_\infty
\lesssim\frac{\sqrt{\log N}}{\sqrt{1-e^{-t}}},\qquad \tau\leq t\leq T-\tau.
\]
These restrictions apply before the fixed support projection in
Section~\ref{form:actor-section}. Under the normalization in
\eqref{eq:cdm-forward}, take
\[
\tau=4N^{-2\alpha/(\alpha\wedge1)},\qquad T=4\alpha\log N.
\]
Use at least
$\lceil N^{d_{\mathcal Y}+d_{\mathcal Z}+12\alpha+4\alpha/(\alpha\wedge1)}\rceil$
independent auxiliary time--Gaussian pairs, shared across the training
observations and independent of them. The times are uniform on
$[\tau,T-\tau]$, and the Gaussian vectors are standard normal.
The nominal reverse grid obeys the sufficient mesh bound in the proof
of Lemma~\ref{aux:conditional-diffusion}. For the projected actor, also take
$C_N\Delta\leq N^{-2\alpha}(\log N)^{19}$, with the uniform Euler error
constant $C_N$ of Lemma~\ref{aux:projected-cdm-sampling}.

For nominal transition estimation, use
$(d_{\mathcal Y},d_{\mathcal Z})=(d_{\mathcal X},d)$,
$N\asymp n_x^{1/(d_{\mathcal X}+d+2\alpha)}$, $T_0=T$, and
$t_{\min}=\tau$. For the actor's separately chosen resolution $N$, use
$(d_{\mathcal Y},d_{\mathcal Z})=(d_{\mathcal A},d_{\mathcal X})$ and
$m\geq N^{d+2\alpha}$. The nominal transition-model and actor score-network
widths are, respectively,
$W_{\mathrm{model}}=\widetilde{\mathcal O}
\left(n_x^{(d_{\mathcal X}+d)/(d_{\mathcal X}+d+2\alpha)}\right),~
W_{\mathrm{actor}}=\widetilde{\mathcal O}
\left(m^{d/(d+2\alpha)}\right).
$
For the actor, Gaussian smoothing establishes the density conditions for
the weighted target, as proved in Appendix~\ref{app:actor-proofs}.

\begin{lemma}[Finite-sample robust critic error]\label{thm:final-critic-rate}
Suppose Assumptions~\ref{ass:cdm-regularity}
and~\ref{ass:fixed-model-coverage} hold.
Use the score classes and sampling choices above, with deterministic
$L_\pi$ and $C\geq1$ independent of $n,m,n_x$.
Fix
$L_0\geq\max\{1,B,L_r+3\gamma BL_p/p_-\}$ independently of the sample sizes.
Take both DNN classes in \eqref{form:critic-fit} to have Lipschitz constants
at most $L_0$ in maximum norm, 
and set the network architecture as
\[
J\asymp\log n,\qquad W\asymp n^{d/(d+2)},\qquad
S\asymp n^{d/(d+2)}(\log n)^3,\qquad M=L_0.
\]
Suppose the on-policy query state density is at least $p_->0$.
Take $m\geq n$.
For positive radii, with $\rho\leq1$ for TV, the minimizer of
\eqref{form:critic-fit} satisfies the following bounds.

\par\smallskip\noindent\textup{(i)} $D=W_1$.
\begin{align*}
\mathbb E\|\widehat Q_k-Q_k\|_{L^2(\mu_k)}^2
&=\widetilde{\mathcal O}\!\left(
\frac{n^{-4/[(d+2)(d_{\mathcal X}+2)]}}{(1-\gamma)^2}\right)\\
&\quad+\widetilde{\mathcal O}\!\left(
\frac{1}{(1-\gamma)^2}
\min\left\{\left(1-\frac\rho{\sqrt{d_{\mathcal X}}}\right)_+^{4/(d_{\mathcal X}+2)},
m^{-2/(d_{\mathcal X}+2)}\right\}\right)\\
&\quad+\widetilde{\mathcal O}\!\left(
\frac{1}{(1-\gamma)^2}
\min\left\{\left(1-\frac\rho{\sqrt{d_{\mathcal X}}}\right)_+^{4/(d_{\mathcal X}+2)},
n_x^{-4\alpha/[(d_{\mathcal X}+d+2\alpha)(d_{\mathcal X}+2)]}\right\}\right).
\end{align*}

\par\smallskip\noindent\textup{(ii)} $D=\mathrm{TV}$.
\begin{align*}
\mathbb E\|\widehat Q_k-Q_k\|_{L^2(\mu_k)}^2
&=\widetilde{\mathcal O}\!\left(
\frac{n^{-4/[(d+2)(d_{\mathcal X}+2)]}}{(1-\gamma)^2}\right)\\
&\quad+\widetilde{\mathcal O}\!\left(
\frac{1}{(1-\gamma)^2}
\min\left\{\rho^{4/(d_{\mathcal X}+2)},(1-\rho)^{4/(d_{\mathcal X}+2)},m^{-2/(d_{\mathcal X}+2)}\right\}\right)\\
&\quad+\widetilde{\mathcal O}\!\left(
\frac{1}{(1-\gamma)^2}
\min\left\{(1-\rho)^{4/(d_{\mathcal X}+2)},
n_x^{-4\alpha/[(d_{\mathcal X}+d+2\alpha)(d_{\mathcal X}+2)]}\right\}\right).
\end{align*}

\par\smallskip\noindent\textup{(iii)} $D=\mathrm{KL}$.
\begin{align*}
\mathbb E\|\widehat Q_k-Q_k\|_{L^2(\mu_k)}^2
&=\widetilde{\mathcal O}\!\left(
\frac{n^{-4/[(d+2)(d_{\mathcal X}+2)]}}{(1-\gamma)^2}\right)+\mathcal O\!\left(
\frac{1}{(1-\gamma)^2}
\left(\frac\rho{\rho+\log m}\right)^{4/(d_{\mathcal X}+2)}\right)
\\
&\quad+\widetilde{\mathcal O}\!\left(
\frac{m^{-1/(d_{\mathcal X}+2)}}{(1-\gamma)^2}\right)+\mathcal O\!\left(
\frac{1}{(1-\gamma)^2}
\left(\frac\rho{\rho+\log n_x}\right)^{4/(d_{\mathcal X}+2)}\right)\\
&\quad+\widetilde{\mathcal O}\!\left(
\frac{n_x^{-2\alpha/[(d_{\mathcal X}+d+2\alpha)(d_{\mathcal X}+2)]}}{(1-\gamma)^2}\right).
\end{align*}

\par\smallskip\noindent\textup{(iv)} $D=\chi^2$.
\[
\mathbb E\|\widehat Q_k-Q_k\|_{L^2(\mu_k)}^2
=\widetilde{\mathcal O}\!\left(\frac{n^{-2/(d+2)}}{(1-\gamma)^2}\right)
+\widetilde{\mathcal O}\!\left(\frac{\min\{1,\rho,\rho^{-2}m^{-1}\}}{(1-\gamma)^2}\right)
+\widetilde{\mathcal O}\!\left(\frac{n_x^{-\alpha/(d_{\mathcal X}+d+2\alpha)}}{(1-\gamma)^2}\right).
\]
\end{lemma}

\begin{lemma}[Weighted conditional diffusion actor]
\label{thm:final-actor-rate}
Suppose
$\pi_k(da\mid x)\geq\epsilon\,da$, where $0<\epsilon<1$,
and $0\leq\widehat Q_k\leq B$. Fix $0<\lambda\leq1$ and use
the weighted actor with
$\eta=\epsilon(1+B/\lambda)^{-1/2}$.
Let $q_{k,\eta}$ be the population conditional density of the
augmented current-policy state--action pairs, with weights
$\exp(\widehat Q_k(X,A)/\lambda)$ evaluated at the original pairs.
For fixed $\alpha>0$, use the support-consistent score class and sampling
choices above with $N\asymp m^{1/(d_{\mathcal A}+d_{\mathcal X}+2\alpha)}$.
The resulting raw folded sampler satisfies
\[
\mathbb E\int_{\mathcal X}
\TV^2\bigl(\widehat\pi_{k+1}(\cdot\mid x),
q_{k,\eta}(\cdot\mid x)\bigr)\,dx
=\widetilde{\mathcal O}\!\left(m^{-2\alpha/(d_{\mathcal A}+d_{\mathcal X}+2\alpha)}\right).
\]
\end{lemma}

\begin{remark}\label{rem:critic-actor-guarantees}
In Lemma~\ref{thm:final-critic-rate}, Bellman Lipschitz regularity controls
the critic approximation. The construction in
Lemma~\ref{aux:compatible-relu-classes} gives squared approximation error
$\lesssim n^{-2/(d+2)}$, while the full-class statistical term is
$S(\log n)^2/n\asymp n^{-2/(d+2)}(\log n)^5$.
Pearson uncertainty bounds adversarial transition densities in $L^2$,
yielding a sharper residual-to-value transfer.
Lemma~\ref{thm:final-critic-rate} is proved in Appendix~\ref{app:critic-rate-proof}.
Lemma~\ref{thm:final-actor-rate} gives the conditional-density estimation rate
$\widetilde{\mathcal O}(m^{-2\alpha/(d_{\mathcal A}+d_{\mathcal X}+2\alpha)})$,
with response dimension $d_{\mathcal A}$ and conditioning dimension
$d_{\mathcal X}$. Its constants may depend on the fixed parameters
$\epsilon,\lambda$; Gaussian augmentation supplies the required regularity
of the weighted target. Lemma~\ref{thm:final-actor-rate} is proved in
Appendix~\ref{app:actor-rate-proof}.
\end{remark}

\paragraph{Finite-sample allocation.}
For critic sample size $n$, fix
\[
\epsilon=n^{-2/[(d+2)(d+3)]}(\log n)^{5/(d+3)},
\qquad
\lambda=\frac{\epsilon}{\log n},
\qquad
K=\left\lceil\frac{2\log n}
{(d+2)(d+3)\log(1/\gamma)}\right\rceil.
\]
These parameters remain fixed during one run. Choose the actor resolution $N$
before training so that $\epsilon^{-d_{\mathcal X}/2}$ times the square root
of the bound in \eqref{eq:actor-general-resolution} is at most $\epsilon$.
Then choose the common sample size
$m\geq\max\{n,\lceil N^{d+2\alpha}\rceil\}$ satisfying
$
\frac{\log(2+L_\pi)}{m}
\leq n^{-2/(d+2)}(\log n)^5.
$
When increasing $m$, keep $N$, the score-network bounds, auxiliary
perturbation budget, time truncation parameters, and reverse grid fixed.
The network and finite sampler have deterministic parameter bounds,
which provide a uniform $L_\pi$ under this allocation.

\begin{theorem}[Last-iterate robust return]\label{robust:global-finite-round}
Suppose the regularity, coverage, and function-class conditions of
Lemmas~\ref{thm:final-critic-rate} and~\ref{thm:final-actor-rate} hold at each
of the $K$ rounds. Use the initial policy and finite-sample allocation
specified above, with fresh queries at each round. Assume the offline query
density is at least $c_{\mathrm{off}}>0$, independent of sample sizes. For positive uncertainty radii, with
$\rho\leq1$ for TV, the returned policy satisfies the following
bounds:
\par\smallskip\noindent\textup{(i)} $D=W_1$.
\begin{align*}
&\mathbb E[\sup_\pi J_D(\pi)-J_D(\pi_K)]\\
&\quad=\widetilde{\mathcal O}\!\left(
\frac{n^{-2/[(d+2)(d+3)]}}{(1-\gamma)^3}\right)\\
&\quad+\widetilde{\mathcal O}\!\left(
\frac{n^{2/[(d+2)^2(d+3)]}}{(1-\gamma)^3}
\min\left\{(1-\rho/\sqrt{d_{\mathcal X}})_+^{2/(d+2)},
m^{-1/(d+2)}\right\}\right)\\
&\quad+\widetilde{\mathcal O}\!\left(
\frac{n^{2/[(d+2)^2(d+3)]}}{(1-\gamma)^3}
\min\left\{
(1-\rho/\sqrt{d_{\mathcal X}})_+^{2/(d+2)},
n^{2d_{\mathcal A}/[(d+2)^2(d+3)]}
n_x^{-2\alpha/[(d_{\mathcal X}+d+2\alpha)(d+2)]}
\right\}\right).
\end{align*}

\par\smallskip\noindent\textup{(ii)} $D=\mathrm{TV}$.
\begin{align*}
&\mathbb E[\sup_\pi J_D(\pi)-J_D(\pi_K)]\\
&\quad=\widetilde{\mathcal O}\!\left(
\frac{n^{-2/[(d+2)(d+3)]}}{(1-\gamma)^3}\right)\\
&\quad+\widetilde{\mathcal O}\!\left(
\frac{n^{2/[(d+2)^2(d+3)]}}{(1-\gamma)^3}
\min\left\{\rho^{2/(d+2)},(1-\rho)^{2/(d+2)},
m^{-1/(d+2)}\right\}\right)\\
&\quad+\widetilde{\mathcal O}\!\left(
\frac{n^{2/[(d+2)^2(d+3)]}}{(1-\gamma)^3}
\min\left\{
(1-\rho)^{2/(d+2)},
n^{2d_{\mathcal A}/[(d+2)^2(d+3)]}
n_x^{-2\alpha/[(d_{\mathcal X}+d+2\alpha)(d+2)]}
\right\}\right).
\end{align*}

\par\smallskip\noindent\textup{(iii)} $D=\mathrm{KL}$.
\begin{align*}
\mathbb E[\sup_\pi J_D(\pi)-J_D(\pi_K)]
&=\widetilde{\mathcal O}\!\left(
\frac{n^{-2/[(d+2)(d+3)]}}{(1-\gamma)^3}\right)+\mathcal O\!\left(
\frac{n^{2/[(d+2)^2(d+3)]}}{(1-\gamma)^3}
\left(\frac{\rho}{\rho+\log m}\right)^{2/(d+2)}
\right)\\
&+\widetilde{\mathcal O}\!\left(
\frac{n^{2/[(d+2)^2(d+3)]}m^{-1/[2(d+2)]}}
{(1-\gamma)^3}\right)\\
&+\mathcal O\!\left(
\frac{n^{2/[(d+2)^2(d+3)]}}{(1-\gamma)^3}
\left(\frac{\rho}{\rho+\log n_x}\right)^{2/(d+2)}
\right)\\
&+\widetilde{\mathcal O}\left(
\frac{n^{2(d_{\mathcal A}+1)/[(d+2)^2(d+3)]}
n_x^{-\alpha/[(d_{\mathcal X}+d+2\alpha)(d+2)]}}
{(1-\gamma)^3}\right).
\end{align*}

\par\smallskip\noindent\textup{(iv)} $D=\chi^2$.
\begin{align*}
\mathbb E[\sup_\pi J_D(\pi)-J_D(\pi_K)]
&=\widetilde{\mathcal O}\!\left(
\frac{n^{-2/[(d+2)(d+3)]}}{(1-\gamma)^3}\right)\\
&+\widetilde{\mathcal O}\!\left(
\frac{n^{2/[(d+2)^2(d+3)]}}{(1-\gamma)^3}
\min\left\{1,\rho^{1/(d+2)},
\rho^{-2/(d+2)}m^{-1/(d+2)}\right\}\right)\\
&+\widetilde{\mathcal O}\!\left(
\frac{n^{(d_{\mathcal A}+2)/[(d+2)^2(d+3)]}
n_x^{-\alpha/[(d_{\mathcal X}+d+2\alpha)(d+2)]}}
{(1-\gamma)^3}\right).
\end{align*}
\end{theorem}

\begin{samepage}
Applying Theorem~\ref{robust:global-finite-round} with the sample allocations
below yields Corollary~\ref{cor:robust-n-rate}, which expresses the return rate
in terms of the critic sample size $n$.

\begin{corollary}[A rate in the critic sample size]
\label{cor:robust-n-rate}
Use the critic architecture of Lemma~\ref{thm:final-critic-rate}.
Under the conditions and actor allocation of
Theorem~\ref{robust:global-finite-round}, fix $\gamma$ and a positive
uncertainty radius $\rho$. For $D\in\{W_1,\mathrm{TV},\chi^2\}$, take
$
n_x \gtrsim 
n^{(d_{\mathcal X}+d+2\alpha)(2d+d_{\mathcal A}+6)/[\alpha(d+2)(d+3)]}.
$
For $D=\mathrm{KL}$, take
$
\min(m,n_x)\gtrsim\exp\!\left(n^{1/(d+2)}\right).
$
Then, for every initial evaluation law $\nu$,
\[
\mathbb E[\sup_\pi J_D(\pi)-J_D(\pi_K)]
=\widetilde{\mathcal O}\!\left(n^{-2/[(d+2)(d+3)]}\right).
\]
\end{corollary}
\end{samepage}

\begin{remark}\label{rem:robust-return-guarantee}
Theorem~\ref{robust:global-finite-round} separates critic estimation and policy
regularization, robust-target integration, and nominal-model estimation,
with dependence on $n,m,n_x$ and $\rho$. Its proof in
Appendix~\ref{app:return-rate-proof} converts integrated residual bounds
into uniform value control under explicit exploration.

Corollary~\ref{cor:robust-n-rate} chooses sample allocations that make the
additional error terms no larger in order than the critic term.
The stated offline budget $n_x$ is polynomial for $W_1$, TV, and Pearson
uncertainty. For KL uncertainty, the exponential allocations for $m,n_x$
control the inverse-logarithmic cutoff terms retained in the theorem.
The actor allocation remains in force, so the resulting rate is measured
in the per-round critic sample size $n$.
\end{remark}

\section{Conclusion}\label{sec:conclusion}
We propose an actor--critic framework for four rectangular transition
uncertainty families on continuous state and action spaces. A fixed
conditional diffusion model estimates nominal transitions, and a second model
learns relative policy improvements through weighted denoising.
ReLU DNN classes control approximation error for the critic and auxiliary function,
while regularized conditional diffusion provides a quantitative policy-improvement
bound. Explicit exploration and robust Bellman contraction yield nonasymptotic
last-iterate return guarantees with explicit dependence on the uncertainty radius,
all three sample sizes, and the model-estimation and integration errors.
Theoretical analysis for dependent data and numerical validation of the
proposed method are left for future work.

\appendix
\section*{Appendix}
In this appendix, we provide the proofs and auxiliary results used in the analysis.
Appendix~\ref{app:inner-proofs} proves the robust dual representations.
Appendices~\ref{app:critic-proofs} and~\ref{app:actor-proofs} establish the critic
and actor bounds, respectively. Appendix~\ref{app:outer-proofs} proves the return
guarantees, and Appendix~\ref{app:auxiliary-results} provides the conditional
diffusion bounds used in these proofs.

\section{Proof of Lemma~\ref{robdual:four}}\label{app:inner-proofs}
We prove Lemma~\ref{robdual:four} below, treating the Wasserstein,
total variation, KL, and Pearson uncertainty sets in turn.

\begin{proof}[Proof of Lemma~\ref{robdual:four}]
For $W_1$, couplings with first marginal $P$ form a compact convex set,
and the diagonal coupling is strictly feasible. Convex separation gives
\[
\inf_{q\in\mathcal U_{W_1}(P;\rho)}\int v\,dq
=\sup_{\lambda\geq0}
\left\{\inf_{\kappa:\,\kappa_1=P}
\int[v(y)+\lambda\|x-y\|_2]\,d\kappa-\lambda\rho\right\}.
\]
A measurable minimizing selection identifies the inner infimum with the
integral of the pointwise minimum, whose values lie in $[0,B]$.
For $\lambda>B/\rho$, the objective is negative, whereas its value
at zero is $\min_{\mathcal X}v\geq0$. Continuity on $[0,B/\rho]$
therefore gives the maximum.

For TV, $(q-P)_-\leq P$ has mass at most $\rho$. Removing a density
$h$ from $P$ and placing its mass at a minimizer of $v$ gives
\[
\inf_{q\in\mathcal U_{\mathrm{TV}}(P;\rho)}\int v\,dq
=\int v\,dP-
\sup_{\substack{0\leq h\leq1\\\int h\,dP\leq\rho}}
\int h(v-\min_{\mathcal X}v)\,dP.
\]
The inequality
$h(v-\min_{\mathcal X}v)\leq hu+(v-\min_{\mathcal X}v-u)_+$
and thresholding $h$ at $v-\min_{\mathcal X}v=u$ yield
\[
\sup_{\substack{0\leq h\leq1\\\int h\,dP\leq\rho}}
\int h(v-\min_{\mathcal X}v)\,dP
=\min_{0\leq u\leq B-\min_{\mathcal X}v}
\left\{\rho u+\int(v-\min_{\mathcal X}v-u)_+\,dP\right\}.
\]
At a threshold atom, take a fractional $h$; if all positive values
have been removed, the remaining budget is unused.
Substituting $t=u+\min_{\mathcal X}v$ gives the TV formula on
$[\min_{\mathcal X}v,B]$. Below $\min_{\mathcal X}v$ its objective is
$(1-\rho)t+\rho\min_{\mathcal X}v$, so extending the domain to $[0,B]$
preserves the maximum.

For KL, $q=P$ is strictly feasible. Convexity of relative entropy and
boundedness of $v$ give a finite, convex, nonincreasing budget-value
function on $(0,\infty)$. A supporting line
at $\rho>0$ has a finite, nonpositive slope; its negative is a
Lagrange multiplier. The supporting-line inequality and weak duality
give strong duality. For $\eta>0$, set
\[
\frac{dq_\eta}{dP}(x)
=\frac{e^{-v(x)/\eta}}{\int e^{-v/\eta}\,dP}.
\]
For every $q\ll P$,
\[
\int v\,dq+\eta\operatorname{KL}(q\Vert P)
=-\eta\log\int e^{-v/\eta}\,dP
+\eta\operatorname{KL}(q\Vert q_\eta),
\]
which identifies the Lagrangian infimum.
For every $\epsilon>0$, conditioning $P$ on
$\{v<\essinf_P v+\epsilon\}$ identifies the
zero-multiplier infimum over finite-entropy laws and gives
\[
0\leq-\eta\log\int e^{-v/\eta}\,dP-\essinf_P v
\leq\epsilon-\eta\log P\{v<\essinf_P v+\epsilon\}.
\]
Letting $\eta\downarrow0$ and then $\epsilon\downarrow0$ gives
the endpoint $\essinf_P v$. Since the objective is at most
$B-\eta\rho$, temperatures larger than $B/\rho$ cannot improve
its nonnegative endpoint value.

For Pearson uncertainty, a feasible $w=dq/dP$ satisfies
$\int w\,dP=1$ and $\int w^2dP\leq1+\rho$. Cauchy--Schwarz gives
\[
\int vw\,dP
=t-\int(t-v)w\,dP
\geq t-\sqrt{(1+\rho)\int(t-v)_+^2\,dP}.
\]
The right side is nonpositive for $t<0$ and negative for
$t>2B(1+\rho)/\rho$, since it is at most
$t-\sqrt{1+\rho}(t-B)$. Its value at zero is zero, so the objective attains
a maximum at some $t^\star$ in the stated interval.
If $\int(t^\star-v)_+^2dP>0$, the maximum is interior and
\[
\sqrt{\int(t^\star-v)_+^2\,dP}
=\sqrt{1+\rho}\int(t^\star-v)_+\,dP.
\]
Then
\[
\frac{dq^\star}{dP}
=\frac{(t^\star-v)_+}{\int(t^\star-v)_+\,dP}
\]
has squared $L^2(P)$ norm $1+\rho$ and attains equality.
Otherwise $t^\star=\essinf_Pv$, and the nonpositive right derivative
$1-\sqrt{(1+\rho)P\{v=t^\star\}}$ implies
$P\{v=t^\star\}\geq(1+\rho)^{-1}$.
Conditioning $P$ on this event is feasible and attains $t^\star$.
\end{proof}

\section{Proof of Lemma~\ref{thm:final-critic-rate}}\label{app:critic-proofs}\label{app:critic-rate-proof}
We first establish the oracle inequality in Lemma~\ref{prop:critic-fast-oracle}.
Lemmas~\ref{form:model-stability}--\ref{aux:compatible-relu-classes} give model
stability, Bellman regularity, and ReLU approximation; these yield the residual
bound in Lemma~\ref{cor:critic-fast-cone-rate}.
Lemmas~\ref{aux:uniform-action-integration}--\ref{aux:kl-center-rate} control
action integration, robust-target sampling, and model error.

\begin{lemma}
\label{prop:critic-fast-oracle}
Conditional on $\mathcal H_k$, let
$(Z_i,Q\mapsto Y_i(Q))$, $i=1,\ldots,n$, be i.i.d random maps with
$Z_i\sim\mu_k$. Suppose $\mathcal Q,\mathcal O$ are compact
$[0,B]$-valued ReLU DNN classes, $B>0$,
$Y_i(Q)\in[0,B]$, and, almost surely,
\[
|Y_i(Q)-Y_i(Q')|\leq\gamma\|Q-Q'\|_\infty
\quad\text{for all }Q,Q'\in\mathcal Q.
\]
Let $\widehat Q_k$ minimize \eqref{form:critic-fit}, and let
$(Z,Y(Q))$ be a generic conditional copy. Suppose the true
operator $\mathcal B_k$ has fixed point $Q_k$ and
is a $\gamma$-contraction in the maximum norm. For every $u>0$ and
$\zeta\in(0,1)$, with conditional probability at least $1-\zeta$,
\begin{align*}
\|\widehat Q_k-\mathcal B_k\widehat Q_k\|_{L^2(\mu_k)}^2
&\lesssim
\inf_{Q\in\mathcal Q}\|Q-Q_k\|_\infty^2
+\sup_{Q\in\mathcal Q}\inf_{O\in\mathcal O}
\|O-\mathcal B_kQ\|_{L^2(\mu_k)}^2\\
&\qquad+\frac{B^2}{n}
\log\!\left[
\frac{2\mathcal N(u,\mathcal Q,\|\cdot\|_\infty)
\mathcal N(u,\mathcal O,\|\cdot\|_\infty)}{\zeta}
\right]+Bu\\
&\qquad+\sup_{Q\in\mathcal Q}\int
\left|\mathbb E[Y(Q)\mid Z=z,\mathcal H_k]
-(\widehat{\mathcal B}_kQ)(z)\right|^2\mu_k(dz)\\
&\qquad+\sup_{Q\in\mathcal Q}\int
\left|(\widehat{\mathcal B}_kQ)(z)
-(\mathcal B_kQ)(z)\right|^2\mu_k(dz).
\end{align*}
The covering numbers are internal maximum-norm covering numbers.
\end{lemma}
\begin{proof}
Define the shared loss
\[
\ell_{Q,O}(Z,Y)
=(Q(Z)-Y(Q))^2-(O(Z)-Y(Q))^2.
\]
Its factorization gives $|\ell_{Q,O}|\leq B^2$ and
\begin{align*}
\operatorname{Var}(\ell_{Q,O}\mid\mathcal H_k)
&\lesssim B^2\int\left[
\bigl(Q(z)-\mathbb E[Y(Q)\mid Z=z,\mathcal H_k]\bigr)^2\right.\\
&\hspace{28mm}\left.+\bigl(O(z)-\mathbb E[Y(Q)\mid Z=z,\mathcal H_k]\bigr)^2
\right]\mu_k(dz).
\end{align*}
Bernstein's and Young's inequalities, followed by a union bound
over internal $u$-covers and Lipschitz extension, give simultaneously
over $Q,O$
\begin{align*}
&\left|\frac1n\sum_{i=1}^n\ell_{Q,O}(Z_i,Y_i)
-\mathbb E[\ell_{Q,O}(Z,Y)\mid\mathcal H_k]\right|\\
&\quad-\frac14\int\left[
\bigl(Q(z)-\mathbb E[Y(Q)\mid Z=z,\mathcal H_k]\bigr)^2
+\bigl(O(z)-\mathbb E[Y(Q)\mid Z=z,\mathcal H_k]\bigr)^2
\right]\mu_k(dz)\\
&\qquad\lesssim\frac{B^2}{n}\log\!\left[
\frac{2\mathcal N(u,\mathcal Q,\|\cdot\|_\infty)
\mathcal N(u,\mathcal O,\|\cdot\|_\infty)}{\zeta}
\right]+Bu.
\end{align*}
The population identity is
\begin{align*}
\mathbb E[\ell_{Q,O}\mid\mathcal H_k]
&=\int\bigl(Q(z)-\mathbb E[Y(Q)\mid Z=z,\mathcal H_k]\bigr)^2\mu_k(dz)\\
&\quad-\int\bigl(O(z)-\mathbb E[Y(Q)\mid Z=z,\mathcal H_k]\bigr)^2\mu_k(dz).
\end{align*}
For any comparator $\bar Q\in\mathcal Q$, the localized bound and
empirical minimization give
\begin{align*}
&\frac34\int\left|\widehat Q_k(z)
-\left.\mathbb E[Y(Q)\mid Z=z,\mathcal H_k]
\right|_{Q=\widehat Q_k}\right|^2\mu_k(dz)\\
&\quad-\frac54\inf_{O\in\mathcal O}\int
\left|O(z)-\left.\mathbb E[Y(Q)\mid Z=z,\mathcal H_k]
\right|_{Q=\widehat Q_k}\right|^2\mu_k(dz)\\
&\quad-\frac54\int
\left|\bar Q(z)-\mathbb E[Y(\bar Q)\mid Z=z,\mathcal H_k]\right|^2\mu_k(dz)\\
&\qquad\lesssim\frac{B^2}{n}\log\!\left[
\frac{2\mathcal N(u,\mathcal Q,\|\cdot\|_\infty)
\mathcal N(u,\mathcal O,\|\cdot\|_\infty)}{\zeta}
\right]+Bu.
\end{align*}
Move the negative terms to the right, minimize over $\bar Q$, and bound
the auxiliary term by the supremum over candidate critics.
Split the conditional-mean discrepancies at
$\widehat{\mathcal B}_kQ$ and $\mathcal B_kQ$. The squared triangle
inequality and
\[
\|Q-\mathcal B_kQ\|_{L^2(\mu_k)}
\leq(1+\gamma)\|Q-Q_k\|_\infty
\]
give the claim.
\end{proof}

\begin{lemma}
\label{form:model-stability}
Let $P,\widehat P\in\mathcal P(\mathcal X)$,
$v\in C(\mathcal X;[0,B])$, and $\rho>0$.
For $D\in\{W_1,\mathrm{TV}\}$, with $\rho\leq1$ for TV,
\[
\left|\inf_{q\in\mathcal U_D(P;\rho)}\int v\,dq
-\inf_{q\in\mathcal U_D(\widehat P;\rho)}\int v\,dq\right|
\leq B\TV(P,\widehat P).
\]
For Wasserstein uncertainty, the same difference is also bounded
by $(B/\rho)W_1(P,\widehat P)$. For Pearson uncertainty,
\[
\left|\inf_{q\in\mathcal U_{\chi^2}(P;\rho)}\int v\,dq
-\inf_{q\in\mathcal U_{\chi^2}(\widehat P;\rho)}\int v\,dq\right|
\lesssim B\sqrt{1+\rho}\sqrt{\TV(P,\widehat P)}.
\]
For KL uncertainty and every $\eta_{\min}>0$,
\[
\left|\inf_{q\in\mathcal U_{\mathrm{KL}}(P;\rho)}\int v\,dq
-\inf_{q\in\mathcal U_{\mathrm{KL}}(\widehat P;\rho)}\int v\,dq\right|
\leq\rho\eta_{\min}
+\eta_{\min}(e^{B/\eta_{\min}}-1)\TV(P,\widehat P).
\]
\end{lemma}
\begin{proof}
For bounded measurable $g$,
\[
\left|\int g\,dP-\int g\,d\widehat P\right|
\leq(\sup g-\inf g)\TV(P,\widehat P).
\]
The integrands in \eqref{form:dual-tv} and \eqref{form:dual-w1}
have oscillation at most $B$. Their common parameter domains
therefore give the TV bounds. The Wasserstein integrands are also
$\eta$-Lipschitz with $\eta\leq B/\rho$; Kantorovich--Rubinstein
duality gives the additional W1 bound.

For $0\leq t\leq2B$, the square-root inequality gives
\[
\left|
\left(\int(t-v)_+^2\,dP\right)^{1/2}
-\left(\int(t-v)_+^2\,d\widehat P\right)^{1/2}
\right|
\lesssim B\sqrt{\TV(P,\widehat P)}.
\]
For $t\geq2B$, cancellation of $t^2$ and division by the sum of
the square roots, each at least $t-B$, bound the same difference
by $\lesssim B\TV(P,\widehat P)\leq B\sqrt{\TV(P,\widehat P)}$.
Taking suprema in \eqref{form:dual-chi} and multiplying by
$\sqrt{1+\rho}$ proves the Pearson bound.

Write \eqref{form:dual-kl} as the equivalent supremum over $\eta>0$.
By H\"older's inequality,
$-\eta\log\int e^{-v/\eta}\,dP$ is nondecreasing in $\eta$.
Restricting this supremum to $\eta\geq\eta_{\min}$ lowers
each robust value by an amount in $[0,\rho\eta_{\min}]$.
The objective is at least $-\rho\eta_{\min}$ at the left endpoint
and at most $B-\rho\eta$, so both restricted suprema may be taken over
$[\eta_{\min},\eta_{\min}+B/\rho]$. On this common interval,
\[
\eta\left|\log\int e^{-v/\eta}\,dP-\log\int e^{-v/\eta}\,d\widehat P\right|
\leq\eta(e^{B/\eta}-1)\TV(P,\widehat P)
\leq\eta_{\min}(e^{B/\eta_{\min}}-1)\TV(P,\widehat P).
\]
The two truncation biases differ by at most $\rho\eta_{\min}$,
giving the KL bound.
\end{proof}

The pointwise bounds are uniform over $v$, so they remain valid
after taking the supremum over candidate continuations before integration.

\begin{lemma}
\label{prop:robust-bellman-regularity-new}
Suppose that the reward is $L_r$-Lipschitz in maximum norm and that
the nominal transition density satisfies
\[
p_0(y\mid z)\ge p_->0,\qquad
|p_0(y\mid z)-p_0(y\mid z')|
\le L_p\|z-z'\|_\infty
\quad\text{for all }y,z,z'.
\]
For every $v\in C(\mathcal X;[0,B])$, the function
\[
z\longmapsto r(z)+\gamma
\inf_{q\in\mathcal U_D(P_0(\cdot\mid z);\rho_D)}\int v\,dq
\]
is Lipschitz in the maximum norm. Sufficient Lipschitz constants are
\[
\begin{cases}
L_r+\gamma BL_p/2,&D=W_1\text{ or }\mathrm{TV},\\[2pt]
L_r+\gamma BL_p/p_-,&D=\mathrm{KL},\\[2pt]
L_r+3\gamma BL_p/p_-,&D=\chi^2.
\end{cases}
\]
The constants are uniform in $v$ and the weakly continuous
policies defining $v(y)=\int Q(y,a)\pi(da\mid y)$.
\end{lemma}
\begin{proof}
For TV and $W_1$, Lemma~\ref{form:model-stability} and
$\TV(P_0(\cdot\mid z),P_0(\cdot\mid z'))
\le L_p\|z-z'\|_\infty/2$ prove the assertion.

For $0<\eta\le B$, the lower density bound gives
\[
\eta\left|\log\int e^{-v/\eta}p_0(y\mid z)\,dy
-\log\int e^{-v/\eta}p_0(y\mid z')\,dy\right|
\le\frac{BL_p}{p_-}\|z-z'\|_\infty.
\]
The integral difference is at most
$(L_p/p_-)\|z-z'\|_\infty$ times either integral; use
$|\log a-\log b|\le |a-b|/\min(a,b)$.
For $\eta\ge B$, cancellation of the constant part of the integrand
and the lower bound $e^{-B/\eta}$ give
\begin{align*}
&\eta\left|\log\int e^{-v/\eta}p_0(y\mid z)\,dy
-\log\int e^{-v/\eta}p_0(y\mid z')\,dy\right|\\
&\quad\le\eta(e^{B/\eta}-1)
\TV(P_0(\cdot\mid z),P_0(\cdot\mid z'))
\le\frac{e-1}{2}BL_p\|z-z'\|_\infty
\le\frac{BL_p}{p_-}\|z-z'\|_\infty.
\end{align*}
Here, $p_-\le1$ because $\mathcal X$ has unit volume. At zero
temperature both centers give the same minimum of $v$; taking
suprema in the KL dual proves the bound.

For Pearson ambiguity with $0<\rho\leq1$ and $0\le t\le2B$,
the density comparison gives
\[
\left|\left(\int(t-v)_+^2p_0(y\mid z)\,dy\right)^{1/2}
-\left(\int(t-v)_+^2p_0(y\mid z')\,dy\right)^{1/2}\right|
\le\frac{2BL_p}{p_-}\|z-z'\|_\infty.
\]
For $t\ge2B$, cancel $t^2$ using
$\int[p_0(y\mid z)-p_0(y\mid z')]\,dy=0$.
The oscillation of $-2tv+v^2$ is at most $B(2t-B)$ and the
square-root sum is at least $2(t-B)$, so
\begin{align*}
&\left|\left(\int(t-v)^2p_0(y\mid z)\,dy\right)^{1/2}
-\left(\int(t-v)^2p_0(y\mid z')\,dy\right)^{1/2}\right|\\
&\quad\le\frac{B(2t-B)}{4(t-B)}L_p\|z-z'\|_\infty
\le\frac34 BL_p\|z-z'\|_\infty.
\end{align*}
Multiplication by $\sqrt{1+\rho}\leq\sqrt2$ gives the constant
$3BL_p/p_-$ for $0<\rho\leq1$.
For $\rho\geq1$, the maximizing dual threshold lies in
$[0,2B(1+\rho)/\rho]\subseteq[0,4B]$. The density bounds give
\begin{align*}
&\left|\left(\int(t-v)_+^2p_0(y\mid z)\,dy\right)^{1/2}
-\left(\int(t-v)_+^2p_0(y\mid z')\,dy\right)^{1/2}\right|\\
&\quad\leq\frac{L_p\|z-z'\|_\infty}{2p_-}
\min\left\{
\left(\int(t-v)_+^2p_0(y\mid z)\,dy\right)^{1/2},
\left(\int(t-v)_+^2p_0(y\mid z')\,dy\right)^{1/2}\right\}.
\end{align*}
Divide the integral-difference bound
$(L_p/p_-)\|z-z'\|_\infty$ times either integral by the
square-root sum; the zero case is immediate. At a maximizing
threshold for the first center, nonnegativity of the dual gives
\[
\sqrt{1+\rho}\left(\int(t-v)_+^2p_0(y\mid z)\,dy\right)^{1/2}\leq t.
\]
At this threshold, the two dual objectives differ by at most
$tL_p\|z-z'\|_\infty/(2p_-)$, hence at most
$2BL_p\|z-z'\|_\infty/p_-$. Interchange the centers and add
the reward.
\end{proof}

\begin{lemma}\label{aux:compatible-relu-classes}
Suppose Assumption~\ref{ass:cdm-regularity} holds and $\pi_k$ is weakly continuous.
Use the DNN architecture and Lipschitz bound in
Lemma~\ref{thm:final-critic-rate}, and put $h=n^{-1/(d+2)}$.
For every $L_0$-Lipschitz $f:[0,1]^d\to[0,B]$,
\[
\inf_{Q\in\mathcal Q}\|Q-f\|_\infty
+\inf_{O\in\mathcal O}\|O-f\|_\infty\lesssim h.
\]
The fitted Bellman residual is $2L_0$-Lipschitz. For $0<u\leq1$,
\[
\log\mathcal N(u,\mathcal Q,\|\cdot\|_\infty)
+\log\mathcal N(u,\mathcal O,\|\cdot\|_\infty)
\lesssim S\left[(J+1)\log\bigl((W+d+1)(M+1)\bigr)
+\log\frac{J+1}{u}\right].
\]
\end{lemma}
\begin{proof}
Choose an $h$-grid $\mathcal V_h\subset[0,1]^d$ with
$|\mathcal V_h|\lesssim h^{-d}$. Then
\[
0\leq\min\!\left(B,\min_{v\in\mathcal V_h}
[f(v)+L_0\|z-v\|_\infty]\right)-f(z)\leq2L_0h.
\]
The lower bound follows from the Lipschitz inequality for $f$; a grid
point within distance $h$ gives the upper bound. Each cone is
$L_0$-Lipschitz in maximum norm, and finite minima and clipping at $B$
preserve this bound.
The identities
\[
|u|=\sigma(u)+\sigma(-u),\qquad
\max(u,v)=\sigma(u-v)+v,\qquad
\min(u,v)=u-\sigma(u-v)
\]
realize the approximation exactly by a ReLU DNN. Coordinate differences
use weights $\pm1$ and biases bounded by one; multiplication by $L_0$
uses a weight $L_0$. The offsets $f(v)$ and clipping threshold have
magnitude at most $B$. Minimum, maximum, and copying operations use
weights $\pm1$. Hence every parameter is bounded by $L_0$, since
$L_0\geq\max\{1,B\}$.
Balanced maximum and minimum trees give depth $\lesssim\log(1/h)$
and width and size $\lesssim h^{-d}=n^{d/(d+2)}$, within the stated budgets.
The output is $[0,B]$-valued and $L_0$-Lipschitz, so the approximation
lies in both DNN classes. Lemma~\ref{prop:robust-bellman-regularity-new} gives
the residual modulus and places $Q_k=\mathcal B_kQ_k$ and every
$\mathcal B_kQ$, $Q\in\mathcal Q$, in the same Lipschitz ball.

There are at most $\exp\{CS\log((J+1)(W+d+1))\}$ sparsity patterns.
On each pattern, the parameter-to-function map has maximum-norm
Lipschitz constant at most
$(J+1)((W+d+1)(M+1))^{J+1}$.
Cover each parameter cube at the corresponding resolution, retaining
one admissible network per intersecting box. This gives the stated
internal covering bound for both DNN classes.
\end{proof}

\begin{lemma}
\label{cor:critic-fast-cone-rate}
Suppose the conditions of Lemma~\ref{prop:critic-fast-oracle}
and Assumption~\ref{ass:cdm-regularity} hold. Use the
ReLU DNN classes $\mathcal Q$ and $\mathcal O$ with the architecture and
Lipschitz bound in Lemma~\ref{thm:final-critic-rate}.
Then, taking expectation over all training and target-generation
randomness,
\begin{align*}
\mathbb E\|\widehat Q_k
-\mathcal B_k\widehat Q_k\|_{L^2(\mu_k)}^2
&\lesssim
n^{-2/(d+2)}(\log n)^5 +\mathbb E\sup_{Q\in\mathcal Q}\int
\left|\mathbb E[Y(Q)\mid Z=z,\mathcal H_k]
-(\widehat{\mathcal B}_kQ)(z)\right|^2\mu_k(dz)\\
&\quad+\mathbb E\sup_{Q\in\mathcal Q}\int
\left|(\widehat{\mathcal B}_kQ)(z)
-(\mathcal B_kQ)(z)\right|^2\mu_k(dz).
\end{align*}
\end{lemma}
\begin{proof}
Put $h=n^{-1/(d+2)}$.
Lemma~\ref{aux:compatible-relu-classes} gives squared approximation
errors $\lesssim h^2$ and
\[
\log\mathcal N(u,\mathcal Q,\|\cdot\|_\infty)
+\log\mathcal N(u,\mathcal O,\|\cdot\|_\infty)
\lesssim S\bigl[(\log n)^2+\log(1/u)\bigr],\qquad 0<u\leq1.
\]
Apply Lemma~\ref{prop:critic-fast-oracle} with $u=h^2$ and
$\zeta=1/n$. Since $S\asymp n^{d/(d+2)}(\log n)^3$,
\[
h^2+\frac{S(\log n)^2}{n}
\lesssim n^{-2/(d+2)}(\log n)^5.
\]
The failure contribution is at most $B^2/n\lesssim h^2$.
Take expectations over $\mathcal H_k$ and retain both discrepancy terms.
\end{proof}

\begin{lemma}\label{aux:uniform-action-integration}
Condition on the preceding fitted history. Suppose
$\mathbb E\operatorname{Lip}(A_k(\cdot,\zeta))^2\leq L_\pi^2$,
with maximum-norm states and Euclidean actions. For the DNN classes in
Lemma~\ref{thm:final-critic-rate} and $m\geq n$,
\[
\mathbb E\sup_{Q\in\mathcal Q}\sup_y
\left|\widetilde V_{k,Q}(y)-\int Q(y,a)\pi_k(da\mid y)\right|^2
\lesssim\left[n^{-2/(d+2)}(\log n)^5
+\frac{\log(2+L_\pi)}m\right].
\]
\end{lemma}
\begin{proof}
Markov's inequality gives
\[
\Pr\!\left(\frac1m\sum_{j=1}^m
\operatorname{Lip}(A_k(\cdot,\zeta_{k,j}))
>m(1+L_\pi)\right)\leq\frac1m.
\]
An internal critic cover of resolution $B/m$ and a state grid of
mesh $B/[m^2L_0(2+2L_\pi)]$ give extension error at most
$4B/m$ on the complementary event. The logarithm of their combined cardinality
is at most
\[
C\left[S\bigl((\log n)^2+\log(2m)\bigr)+\log(2+L_\pi)\right].
\]
Hoeffding's inequality and a union bound over these deterministic pairs give
failure probability $1/m$, before intersecting with the Markov
event. Squaring and using the exceptional-event bound $B^2$ gives
\[
C\frac{S\bigl((\log n)^2+\log(2m)\bigr)+\log(2+L_\pi)}m.
\]
The failure contribution $2B^2/m$ is absorbed. Use
$\log(2m)/m\lesssim\log(2n)/n$ for $m\geq n$ and
$S\asymp n^{d/(d+2)}(\log n)^3$.
\end{proof}

\begin{lemma}
\label{aux:target-sampling}
Let $\mathcal X=[0,1]^{d_{\mathcal X}}$, $v\in C(\mathcal X;[0,B])$,
and $P_m=m^{-1}\sum_{i=1}^m\delta_{X_i}$, where the $X_i$ are independent
with law $P$ and $m\geq1$. For positive uncertainty radii, with
$0<\rho\leq1$ in the TV case,
\[
\begin{aligned}
&\left|\mathbb E\left[\inf_{q\in\mathcal U_D(P_m;\rho)}\int v\,dq\right]
-\inf_{q\in\mathcal U_D(P;\rho)}\int v\,dq\right|^2
\\
&\qquad\lesssim
\begin{cases}
B^2\displaystyle\min\left\{1,
\frac{\log[2(2+m\sqrt{d_{\mathcal X}}/\rho)]}{m}\right\},&D=W_1,\\[2mm]
B^2\min\{\rho^2,(1-\rho)^2,1/m\},&D=\mathrm{TV},\\[1mm]
B^2\displaystyle\min\left\{1,\rho,
\frac{(1+\rho)^3\log[2(m+1)]}{\rho^2m}\right\},&D=\chi^2,\\[2mm]
\displaystyle\frac{B^2[\rho^2+e^{2\rho}/\sqrt m]}
{[\rho+\log(m+1)]^2},&D=\mathrm{KL}.
\end{cases}
\end{aligned}
\]
If $v$ is $H$-Lipschitz in the Euclidean norm, the Wasserstein bound
can also be replaced by its minimum with $H^2\rho^2$. The KL bound
also holds with the left side replaced by
\[
\mathbb E\left|
\inf_{q\in\mathcal U_{\mathrm{KL}}(P_m;\rho)}\int v\,dq
-\inf_{q\in\mathcal U_{\mathrm{KL}}(P;\rho)}\int v\,dq
\right|^2.
\]
The bounds are uniform in $P,v$ and apply conditionally
with a fixed learned kernel and action bank.
\end{lemma}

\begin{proof}
The Wasserstein and TV duals are suprema of affine functions of $P$; the
Pearson dual is convex in $P$ by concavity of the square root.
Jensen's inequality and feasibility of the center give
\[
0\leq
\mathbb E\left[\inf_{q\in\mathcal U_D(P_m;\rho)}\int v\,dq\right]
-\inf_{q\in\mathcal U_D(P;\rho)}\int v\,dq
\leq\int v\,dP-\inf_{q\in\mathcal U_D(P;\rho)}\int v\,dq.
\]
Every robust value lies in $[0,B]$.

For $W_1$, put
\[
\psi_u(x)=\min_{y\in\mathcal X}[v(y)+u\|x-y\|_2],
\qquad 0\leq u\leq B/\rho.
\]
These functions take values in $[0,B]$ and satisfy
$\|\psi_u-\psi_{u'}\|_\infty\leq\sqrt{d_{\mathcal X}}|u-u'|$.
Let $M=\lceil m\sqrt{d_{\mathcal X}}/\rho\rceil$ and
$u_j=jB/(\rho M)$, $0\leq j\leq M$.
Hoeffding's exponential bound on this grid yields
\[
\begin{split}
\mathbb E\sup_{0\leq u\leq B/\rho}
\left|\int\psi_u\,d(P_m-P)\right|
&\leq\frac{2B}{m}
+B\sqrt{\frac{\log[2(M+1)]}{2m}},\\
\left(\mathbb E\sup_{0\leq u\leq B/\rho}
\left|\int\psi_u\,d(P_m-P)\right|\right)^2
&\leq\frac{8B^2}{m^2}
+\frac{B^2\log[2(M+1)]}{m}.
\end{split}
\]
The empirical supremum bounds the dual difference; use
$M+1\leq2+m\sqrt{d_{\mathcal X}}/\rho$ to obtain the stated rate.
For $H$-Lipschitz $v$, Kantorovich--Rubinstein duality gives
$\int v\,dP-\inf_{q\in\mathcal U_{W_1}(P;\rho)}\int v\,dq\leq H\rho$.

For the distribution functions $F,F_m$ of $v(X)$ and its empirical
law, the Dvoretzky--Kiefer--Wolfowitz inequality and tail integration give
\[
\mathbb P\left(\sup_s|F_m(s)-F(s)|>t\right)\leq2e^{-2mt^2},
\qquad
\mathbb E\sup_s|F_m(s)-F(s)|^2\leq\frac1m.
\]
For any monotone function $f$ on $[0,B]$, integration by parts gives
\[
\left|\int f(v)\,d(P_m-P)\right|
\leq |f(B)-f(0)|\sup_s|F_m(s)-F(s)|.
\]
For TV, $f(z)=\min(z,u)$ has variation at most $B$, giving
squared conditional-mean error $B^2/m$. The $\rho^2B^2$ bound follows from
\[
\int v\,dP-\rho B
\leq\inf_{q\in\mathcal U_{\mathrm{TV}}(P;\rho)}\int v\,dq
\leq\int v\,dP.
\]
For a minimizer $y$, the feasible law $(1-\rho)P+\rho\delta_y$
places both robust values in
$[\min_{\mathcal X}v,\min_{\mathcal X}v+(1-\rho)B]$,
giving the remaining cap.

For Pearson uncertainty, let $u_{\max}=2B(1+\rho)/\rho$ and
$g_u=(u-v)_+$, $0\leq u\leq u_{\max}$. Since
$0\leq g_u^2\leq u_{\max}^2$ and
$\operatorname{Var}_P(g_u^2)\leq u_{\max}^2\int g_u^2\,dP$,
Bernstein's inequality gives, with probability at least $1-2e^{-t}$,
\[
\left|\int g_u^2\,d(P_m-P)\right|
\leq u_{\max}\sqrt{\frac{2t}{m}\int g_u^2\,dP}
+\frac{2u_{\max}^2t}{3m}.
\]
Solving the square-root inequalities gives
\[
\left|\sqrt{\int g_u^2\,dP_m}-\sqrt{\int g_u^2\,dP}\right|
\leq2u_{\max}\sqrt{t/m}.
\]
Both square roots are 1-Lipschitz in $u$. A union bound on
$u_j=ju_{\max}/m$, $0\leq j\leq m$, gives, with probability
at least $1-e^{-s}$,
\[
\sup_{0\leq u\leq u_{\max}}
\left|\sqrt{\int g_u^2\,dP_m}-\sqrt{\int g_u^2\,dP}\right|
\leq2u_{\max}\sqrt{\frac{\log[2(m+1)]+s}{m}}
+\frac{2u_{\max}}m.
\]
Squaring and integrating this tail yields
\[
\begin{split}
\mathbb E\sup_{0\leq u\leq u_{\max}}
\left|\sqrt{\int g_u^2\,dP_m}-\sqrt{\int g_u^2\,dP}\right|^2
&\leq\frac{8u_{\max}^2[\log[2(m+1)]+1]}m
+\frac{8u_{\max}^2}{m^2}\\
&\leq\frac{20u_{\max}^2\log[2(m+1)]}{m}.
\end{split}
\]
The Pearson dual gives the corresponding bound.
Cauchy--Schwarz also gives
\[
\int v\,dP-\inf_{q\in\mathcal U_{\chi^2}(P;\rho)}\int v\,dq
\leq\sqrt\rho\,\sqrt{\operatorname{Var}_P(v)}
\leq\frac{B\sqrt\rho}{2},
\]
which proves the additional $\rho B^2/4$ bound.

For KL, H\"older's inequality implies that
$u\mapsto-u\log\int e^{-v/u}\,dP$ is nondecreasing. Restricting the
dual to $u\geq\eta>0$ therefore gives
\[
0\leq\inf_{q\in\mathcal U_{\mathrm{KL}}(P;\rho)}\int v\,dq
-\sup_{u\geq\eta}\left[-u\log\int e^{-v/u}\,dP-u\rho\right]
\leq\rho\eta.
\]
The two truncation errors lie in the same interval, so their
difference is at most $\rho\eta$. Since the exponential integrals
are at least $e^{-B/u}$, the variation bound gives
\[
u\left|\log\int e^{-v/u}\,dP_m-\log\int e^{-v/u}\,dP\right|
\leq u(e^{B/u}-1)\sup_s|F_m(s)-F(s)|.
\]
The coefficient $u(e^{B/u}-1)$ decreases in $u$. Consequently,
$$\mathbb E\left|
\inf_{q\in\mathcal U_{\mathrm{KL}}(P_m;\rho)}\int v\,dq
-\inf_{q\in\mathcal U_{\mathrm{KL}}(P;\rho)}\int v\,dq
\right|^2 \leq2\rho^2\eta^2+
\frac{2\eta^2(e^{B/\eta}-1)^2}{m}.
$$
Take $\eta=B/[\rho+\tfrac14\log(m+1)]$ and use
$\sqrt{m+1}/m\leq\sqrt2/\sqrt m$ and $2\sqrt2<3$.
Jensen's inequality gives the squared conditional-mean bound.
\end{proof}

\begin{lemma}
\label{aux:kl-center-rate}
Under Assumption~\ref{ass:fixed-model-coverage},
\eqref{eq:conditional-model-rate} gives, for
$\alpha>0$, the unconditional bound
\[
\mathbb E\int
\TV^2\!\left(\widehat P(\cdot\mid x,a),P_0(\cdot\mid x,a)\right)
\mu_k(dx,da)
\leq CC_{\mathrm{CDM}}
n_x^{-2\alpha/(d_{\mathcal X}+d+2\alpha)}
(\log n_x)^{19}.
\]
For every $\rho>0$ and every family
$v_{x,a}\in C(\mathcal X;[0,B])$,
\[
\begin{split}
&\mathbb E\int
\left|\inf_{q\in\mathcal U_{\mathrm{KL}}(\widehat P(\cdot\mid x,a);\rho)}
\int v_{x,a}\,dq
-\inf_{q\in\mathcal U_{\mathrm{KL}}(P_0(\cdot\mid x,a);\rho)}
\int v_{x,a}\,dq\right|^2\mu_k(dx,da)\\
&\qquad\leq
\frac{2B^2\left[\rho^2+CC_{\mathrm{CDM}}e^{2\rho}
n_x^{-\alpha/(d_{\mathcal X}+d+2\alpha)}
(\log n_x)^{19}\right]}
{\left[\rho+\dfrac{\alpha\log n_x}
{2(d_{\mathcal X}+d+2\alpha)}\right]^2}.
\end{split}
\]
The expectation includes the offline data and fitted history; the
bound is uniform over adaptive families.
\end{lemma}

\begin{proof}
For any two probability laws $P,\widehat P$ and $\eta>0$, the KL
truncation argument above and the bounded-function TV inequality give
\[
\left|\inf_{q\in\mathcal U_{\mathrm{KL}}(\widehat P;\rho)}\int v\,dq
-\inf_{q\in\mathcal U_{\mathrm{KL}}(P;\rho)}\int v\,dq\right|
\leq\rho\eta+\eta(e^{B/\eta}-1)\TV(\widehat P,P).
\]
Square and integrate under $\mu_k$. Pathwise coverage in
Appendix~\ref{app:cdm-import}, followed by unconditional expectation,
applies also when $\mu_k$ depends on the fitted model. Choosing
\[
\eta=\frac{B}{\rho+\alpha\log n_x/
[2(d_{\mathcal X}+d+2\alpha)]}
\]
gives the result because
\[
(e^{B/\eta}-1)^2\leq e^{2\rho}
n_x^{\alpha/(d_{\mathcal X}+d+2\alpha)}.
\]
The pointwise bound is independent of $v$, so adaptive families
satisfy it as well.
\end{proof}

Combining the residual bound in Lemma~\ref{cor:critic-fast-cone-rate}
with the stability, sampling, and model-error bounds in
Lemmas~\ref{form:model-stability} and~\ref{aux:uniform-action-integration}--\ref{aux:kl-center-rate},
we now prove Lemma~\ref{thm:final-critic-rate}.

\begin{proof}[Proof of Lemma~\ref{thm:final-critic-rate}]
Apply Lemma~\ref{cor:critic-fast-cone-rate}. Robust expectations are
nonexpansive in the continuation, so Lemma~\ref{aux:uniform-action-integration}
controls the action-bank error; Lemma~\ref{aux:target-sampling} controls the
remaining target bias. For $W_1$,
$\operatorname{Lip}(\widetilde V_{k,Q})\leq
L_0[1+m^{-1}\sum_j\operatorname{Lip}(A_k(\cdot,\zeta_{k,j}))]$,
with expected square at most $C(1+L_\pi)^2$.
If $\rho\leq[(1+L_\pi)\sqrt m]^{-1}$, the Lipschitz radius cap gives
expected squared bias $\lesssim m^{-1}$; otherwise the expected scalar-grid
logarithm is $\lesssim\log(2m)+\log(2+L_\pi)$.
Thus the expected squared bias is $\lesssim[\log(2m)+\log(2+L_\pi)]/m$; the second term
is absorbed for fixed $L_\pi$ and $m\geq n$, or under the finite-sample allocation.
Mixing a center with a global minimizer with weight
$\min\{1,\rho/\sqrt{d_{\mathcal X}}\}$ bounds both squared successor-sampling bias and
center error by $B^2(1-\rho/\sqrt{d_{\mathcal X}})_+^2$.
The analogous TV cap is $B^2(1-\rho)^2$; the remaining radius minima
follow from Lemma~\ref{aux:target-sampling}.

For model error, apply Lemma~\ref{form:model-stability} and
\eqref{eq:conditional-model-rate}, using coverage pathwise before averaging
over the fitted model. Lemma~\ref{aux:kl-center-rate} gives the KL terms.
These bounds are uniform over continuations.

The common-noise bound gives
$W_1(\pi_k(\cdot\mid x),\pi_k(\cdot\mid x'))\leq L_\pi\|x-x'\|_\infty$.
Together with Lemma~\ref{aux:compatible-relu-classes}, this makes the
integral inside the following supremum $2L_0(1+L_\pi)$-Lipschitz.
Integrating over a ball about its maximizer, using the state-density
lower bound and Jensen's inequality, yields
\begin{align*}
&\mathbb E\sup_x\left[\int
|\widehat Q_k(x,a)-(\mathcal B_k\widehat Q_k)(x,a)|
\pi_k(da\mid x)\right]^2\\
&\quad\lesssim
(1+L_\pi)^{2d_{\mathcal X}/(d_{\mathcal X}+2)}
\left(\mathbb E\|\widehat Q_k-\mathcal B_k\widehat Q_k
\|_{L^2(\mu_k)}^2\right)^{2/(d_{\mathcal X}+2)}.
\end{align*}
The ball radius is proportional to the maximum divided by
$B+2L_0(1+L_\pi)$; the volume of its intersection with the state cube is bounded below by
the radius to the power $d_{\mathcal X}$.
Iterating the Bellman difference-of-infima inequality and applying
Minkowski's inequality give
\[
\|\widehat Q_k-Q_k\|_{L^2(\mu_k)}
\leq\|\widehat Q_k-\mathcal B_k\widehat Q_k\|_{L^2(\mu_k)}
+\frac\gamma{1-\gamma}\sup_x\int
|\widehat Q_k(x,a)-(\mathcal B_k\widehat Q_k)(x,a)|
\pi_k(da\mid x).
\]
Squaring, applying the interpolation bound, and using boundedness by $B$ give
\[
\mathbb E\|\widehat Q_k-Q_k\|_{L^2(\mu_k)}^2
\lesssim\frac{(1+L_\pi)^{2d_{\mathcal X}/(d_{\mathcal X}+2)}}{(1-\gamma)^2}
\left(\mathbb E\|\widehat Q_k-\mathcal B_k\widehat Q_k
\|_{L^2(\mu_k)}^2\right)^{2/(d_{\mathcal X}+2)}.
\]
Substituting the residual bounds proves the first three cases.

For Pearson uncertainty, every admissible next-state density satisfies
\[
\int q(y\mid x,a)^2\,dy
\leq p_+\int\left(\frac{dq}{dP_0}\right)^2dP_0
\leq p_+(1+\rho).
\]
Cauchy--Schwarz, Jensen's inequality, and the query-state density lower bound imply
\begin{align*}
&\sup_{x,a}\sup_{q\in\mathcal U_{\chi^2}(x,a)}
\int\!\int|\widehat Q_k(y,b)-(\mathcal B_k\widehat Q_k)(y,b)|
\pi_k(db\mid y)q(dy)\\
&\quad\leq\sqrt{\frac{p_+(1+\rho)}{p_-}}
\|\widehat Q_k-\mathcal B_k\widehat Q_k\|_{L^2(\mu_k)}.
\end{align*}
Iterating the Bellman difference-of-infima inequality and summing the
geometric series, with the bounded terminal remainder vanishing, give
\[
\|\widehat Q_k-Q_k\|_{L^2(\mu_k)}
\leq\left[1+\frac\gamma{1-\gamma}
\sqrt{\frac{p_+(1+\rho)}{p_-}}\right]
\|\widehat Q_k-\mathcal B_k\widehat Q_k\|_{L^2(\mu_k)}.
\]
Squaring and substituting the Pearson residual bounds proves the fourth case.
\end{proof}

\section{Proof of Lemma~\ref{thm:final-actor-rate}}\label{app:actor-proofs}\label{app:weighted-score}\label{app:actor-rate-proof}
We first bound the projected sampler's time-discretization error in
Lemma~\ref{aux:projected-cdm-sampling}. We then combine this bound with
weighted score estimation to prove Lemma~\ref{thm:final-actor-rate}.

\begin{lemma}
\label{aux:projected-cdm-sampling}
Fix the network resolution, architecture, parameter bounds, and positive
cutoff $\tau$. Conditional on a fitted projected score, let $Y$ solve
the actor's learned reverse SDE with initial law $N(0,I_{d_{\mathcal A}})$, and let
$Y^\Delta$ be its continuous Euler interpolation on a deterministic grid
with maximum step $\Delta\leq1$. There is a deterministic finite constant
$C_N$, uniform over fitted histories, networks in the specified class,
conditioning values, and all such grids, such that
\[
\TV^2\!\left(\mathcal L(Y_{T-2\tau}),
\mathcal L(Y^\Delta_{T-2\tau})\right)\leq C_N\Delta.
\]
\end{lemma}
\begin{proof}
Set $b(v,y,z)=y/2+\widehat s(T-\tau-v,y,z)$.
The fixed network bounds, nonexpansive projection, and $\tau>0$ give
$\|b(v,y,z)\|_2\leq C_N(1+\|y\|_2)$ and, uniformly in $z$,
\[
\|b(v,y,z)-b(w,y',z)\|_2
\leq C_N\bigl[\|y-y'\|_2+|v-w|(1+\|y'\|_2)\bigr].
\]
Gronwall's inequality and the Brownian maximal inequality imply
\[
\mathbb E\sup_{v\leq T-2\tau}\|Y_v\|_2^2\leq C_N,\qquad
\mathbb E\|Y_v-Y_w\|_2^2\leq C_N|v-w|
\quad\text{if }|v-w|\leq1.
\]
Let $\mathbb P,\mathbb P^\Delta$ denote the exact and Euler path laws,
and let $\lfloor v\rfloor_\Delta$ be the preceding grid point.
Under $\mathbb P^\Delta$, use $Y$ as the coordinate path and stop
the Brownian drift change
\[
b(v,Y_v,z)-b(\lfloor v\rfloor_\Delta,
Y_{\lfloor v\rfloor_\Delta},z)
\]
at $\sigma_R=\inf\{v\geq0:\|Y_v\|_2\geq R\}\wedge(T-2\tau)$.
Its bounded stopped integrand defines a likelihood $Z_R$ and
$d\mathbb Q_R=Z_R\,d\mathbb P^\Delta$.
The initial Gaussian law is unchanged. Under $\mathbb Q_R$, the
drift is exact before $\sigma_R$ and Euler afterwards; the same
linear growth bound gives
$\sup_R\mathbb E_{\mathbb Q_R}\sup_v\|Y_v\|_2^2\leq C_N$.
The likelihood entropy therefore satisfies
\[
\sup_R\mathbb E_{\mathbb P^\Delta}[Z_R\log Z_R]
\leq C_N\left(1+\sup_R\mathbb E_{\mathbb Q_R}
\sup_v\|Y_v\|_2^2\right)\leq C_N.
\]
Since $\sigma_R\uparrow T-2\tau$ almost surely, uniform integrability
gives $Z_R\to Z_\infty$ in $L^1$ with
$\mathbb E_{\mathbb P^\Delta}Z_\infty=1$.
Uniqueness in law identifies the changed law as $\mathbb P$
\citep{KlebanerLiptser2014}. Girsanov's identity and the increment
bound yield
\[
\KL(\mathbb P\|\mathbb P^\Delta)
\lesssim\mathbb E\int_0^{T-2\tau}
\bigl\|b(v,Y_v,z)-b(\lfloor v\rfloor_\Delta,
Y_{\lfloor v\rfloor_\Delta},z)\bigr\|_2^2\,dv
\leq C_N\Delta.
\]
Pinsker's inequality and data processing under terminal evaluation
and folding prove the result.
\end{proof}
Using the discretization bound in Lemma~\ref{aux:projected-cdm-sampling},
we now prove Lemma~\ref{thm:final-actor-rate} by combining weighted score
estimation with reverse-diffusion sampling.

\begin{proof}[Proof of Lemma~\ref{thm:final-actor-rate}]
For $\pi_k(da\mid x)\geq\epsilon\,da$, $0<\epsilon<1$,
$0\leq\widehat Q_k\leq B$, $0<\lambda\leq1$, and
$\eta=\epsilon(1+B/\lambda)^{-1/2}$, the score class and sampling
choices of Section~\ref{sec:theory} give, whenever
$m\geq\lceil N^{d_{\mathcal A}+d_{\mathcal X}+2\alpha}\rceil$,
\begin{equation}\label{eq:actor-general-resolution}
\mathbb E\int_{\mathcal X}
\TV^2\bigl(\widehat\pi_{k+1}(\cdot\mid x),
q_{k,\eta}(\cdot\mid x)\bigr)\,dx
\lesssim N^{-2\alpha}(\log N)^{19}.
\end{equation}
The bound is uniform over fitted histories.
The reflected Gaussian kernel $K_{r,q}(z,u)$ of
$\mathsf F_q(u+rG)$ is symmetric and bistochastic. For $0<r\leq1$,
\[
\sup_{z,u}K_{r,q}(z,u)\leq C_qr^{-q},\qquad
\sup_z\int|\partial_z^\nu K_{r,q}(z,u)|\,du
\leq C_{q,\nu}r^{-|\nu|}.
\]
The unweighted augmented state is uniform, and the weighted
conditional density is
\[
q_{k,\eta}(a\mid x)
=\frac{\displaystyle\int\!\!\int
K_{\eta,d_{\mathcal X}}(x,u)K_{\eta,d_{\mathcal A}}(a,b)
e^{\widehat Q_k(u,b)/\lambda}\pi_k(db\mid u)\,du}
{\displaystyle\int\!\!\int
K_{\eta,d_{\mathcal X}}(x,u)
e^{\widehat Q_k(u,b)/\lambda}\pi_k(db\mid u)\,du}.
\]
The denominator lies in $[1,e^{B/\lambda}]$. The exploration floor gives the lower bound, and the normalized
mixture of response kernels gives the upper bound:
\[
\epsilon e^{-B/\lambda}\leq q_{k,\eta}(a\mid x)
\leq C_{d_{\mathcal A}}\eta^{-d_{\mathcal A}}.
\]
Differentiating the quotient and Gaussian kernels gives, for every
finite multi-index $\nu$,
\[
\sup_{a,x}|\partial^\nu q_{k,\eta}(a\mid x)|
\leq C_{\nu,d_{\mathcal X},d_{\mathcal A}}
e^{(|\nu|+1)B/\lambda}\eta^{-(d_{\mathcal A}+|\nu|)}.
\]
The reflected series extends smoothly to $[-1,1]^{d_{\mathcal A}}$.
These bounds verify the density conditions in
Lemma~\ref{aux:conditional-diffusion} uniformly over fitted histories.

For $y=e^{-t/2}a+\sqrt{1-e^{-t}}G$, $a\in[0,1]^{d_{\mathcal A}}$,
the denoising label $-G/\sqrt{1-e^{-t}}$ belongs to
\[
\left\{\frac{e^{-t/2}b-y}{1-e^{-t}}:
b\in[0,1]^{d_{\mathcal A}}\right\}.
\]
Projection onto this set fixes the true score, reduces approximation
and denoising errors, and contracts score distances. The absolute difference between two projected squared losses is
bounded by the distance between the reference scores times
\[
\left\|s_1(t,y,x)+\frac{G}{\sqrt{1-e^{-t}}}\right\|_2
+\left\|s_2(t,y,x)+\frac{G}{\sqrt{1-e^{-t}}}\right\|_2.
\]
The reference-class bound in Lemma~\ref{aux:conditional-diffusion}
and the prescribed time truncation parameters give
\[
\frac1{T-2\tau}\int_\tau^{T-\tau}
\frac{dt}{1-e^{-t}}\lesssim1.
\]
Thus the projected loss, averaged over time and noise, is at most
$C\log N$, and the difference between losses for two reference scores
is at most
$C\sqrt{\log N}\|s_1-s_2\|_\infty$.
Let $s^\star$ be the true score. By the upper and lower density bounds, the posterior response law
is comparable to a Gaussian restricted to the scaled response cube. Truncated Gaussian moments therefore give,
uniformly over $(a,x)\in\mathcal A\times\mathcal X$,
\[
\mathbb E_G\left\|
s^\star\!\left(t,e^{-t/2}a+\sqrt{1-e^{-t}}G,x\right)
+\frac{G}{\sqrt{1-e^{-t}}}\right\|_2^2
\lesssim\frac1{1-e^{-t}}.
\]
For each augmented observation $(a,x)$, define its denoising loss by
\[
\ell_s(a,x)=\frac1{T-2\tau}\int_\tau^{T-\tau}\mathbb E_G
\left\|s\!\left(t,e^{-t/2}a+\sqrt{1-e^{-t}}G,x\right)
+\frac{G}{\sqrt{1-e^{-t}}}\right\|_2^2\,dt.
\]
Let $(X,A)$ be an independent copy of an original training pair,
with independent augmentations $(X^\eta,A^\eta)$ and Gaussian $G$, and set
\[
w=e^{\widehat Q_k(X,A)/\lambda},\qquad
Y_t=e^{-t/2}A^\eta+\sqrt{1-e^{-t}}G.
\]
Conditionally on $(A^\eta,X^\eta)$, factorization and Cauchy--Schwarz
for the time and noise averages give
\begin{align*}
\bigl(\ell_s(A^\eta,X^\eta)-\ell_{s^\star}(A^\eta,X^\eta)\bigr)^2\lesssim
\frac{\log N}{T-2\tau}\int_\tau^{T-\tau}\mathbb E_G
\|s(t,Y_t,X^\eta)-s^\star(t,Y_t,X^\eta)\|_2^2\,dt.
\end{align*}
The other squared factor is bounded by $C\log N$ using the
reference-score envelope, the posterior moment bound, and the fact
that projection decreases denoising loss. Multiply by
$w^2\leq e^{B/\lambda}w$ and take expectations. The score identity
makes the weighted excess loss equal to the weighted squared score error, so
\begin{align*}
&\mathbb E\left[w^2
\bigl(\ell_s(A^\eta,X^\eta)-\ell_{s^\star}(A^\eta,X^\eta)\bigr)^2\right]\lesssim\log N\,
\mathbb E\left[w
\bigl(\ell_s(A^\eta,X^\eta)-\ell_{s^\star}(A^\eta,X^\eta)\bigr)\right].
\end{align*}
The absolute excess loss is bounded by $\lesssim\log N$.
The normalized weighted state density lies in
$[e^{-B/\lambda},e^{B/\lambda}]$ relative to the uniform density.
Bernstein's inequality and the reference-class covers therefore give
\begin{align*}
&\mathbb E\left[\frac{w}{T-2\tau}\int_\tau^{T-\tau}
\|\widehat s_{k+1}(t,Y_t,X^\eta)-s^\star(t,Y_t,X^\eta)\|_2^2\,dt\right]\\
&\quad\lesssim 
\inf_s\mathbb E\left[\frac{w}{T-2\tau}\int_\tau^{T-\tau}
\|s(t,Y_t,X^\eta)-s^\star(t,Y_t,X^\eta)\|_2^2\,dt\right]\\
&\qquad+\frac{N^d(\log N)^{14}[(\log N)^4+\log m]}m
+N^{-2\alpha}(\log N)^{18}.
\end{align*}
The infimum is over the fixed projected score class.
The Gaussian truncation and covering argument of
\citet[Lemma~4.3]{Chen2026} gives the auxiliary-perturbation term under
the prescribed budget: weights are bounded, and the fixed projection
contracts score distances and decreases squared denoising losses.
Projection does not increase approximation error, so
\citet[Lemma~4.2]{Chen2026} and
$m\geq\lceil N^{d+2\alpha}\rceil$ bound the weighted score error by
$\lesssim N^{-2\alpha}(\log N)^{18}$.

The score identity \eqref{eq:cdm-score} gives linear growth and
finite spatial Lipschitz bounds for the true reverse drift on the
truncated interval. The localization in
Lemma~\ref{aux:projected-cdm-sampling} therefore also justifies its
Girsanov comparison with the learned continuous diffusion, with
both processes initialized from the forward marginal at time $T-\tau$.
The score-to-law, Gaussian-initialization, and truncation bounds
invoked in Lemma~\ref{aux:conditional-diffusion} give
$C N^{-2\alpha}(\log N)^{19}$.
Lemma~\ref{aux:projected-cdm-sampling} adds $C_N\Delta$ of the same
order under the prescribed mesh. Folding cannot increase TV, proving
\eqref{eq:actor-general-resolution}. Taking
$N=\lfloor m^{1/(d_{\mathcal A}+d_{\mathcal X}+2\alpha)}\rfloor$
yields Lemma~\ref{thm:final-actor-rate}.
\end{proof}

\section{Proof of Theorem~\ref{robust:global-finite-round} and Corollary~\ref{cor:robust-n-rate}}\label{app:outer-proofs}\label{app:return-rate-proof}
We first establish the conditional optimization and Bellman properties in
Lemmas~\ref{lem:conditional-continuity} and~\ref{prop:rsfac-bellman}.
Lemmas~\ref{aux:approximate-robust-pi} and~\ref{aux:uniform-residual-control}
give approximate policy improvement and residual-to-value control.
Lemmas~\ref{prop:weighted-actor-greedy} and~\ref{app:cdm-density} provide the
greedy-error and sampler-coverage bounds used in
Theorem~\ref{robust:global-finite-round}.

\begin{lemma}\label{lem:conditional-continuity}
Under Assumption~\ref{ass:cdm-regularity}, each conditional
uncertainty correspondence $z\mapsto\mathcal U_D(P_0(\cdot\mid z);\rho_D)$
is nonempty, compact-valued, and continuous in the weak topology.
For every $v\in C(\mathcal X)$, the minimum of $\int v\,dq$ over
this correspondence is continuous in $z$, and a Borel minimizing kernel exists.
\end{lemma}
\begin{proof}
The nominal laws are TV-continuous. On the compact metric space
$\mathcal P(\mathcal X)$, $W_1$ is continuous, and TV, KL and $\chi^2$
are jointly lower semicontinuous. For KL and $\chi^2$, the variational representations are
\[
\operatorname{KL}(q\Vert P)
=\sup_{f\in C(\mathcal X)}
\left(\int f\,dq-\log\int e^f\,dP\right),
\]
\[
\chi^2(q\Vert P)
=\sup_{f\in C(\mathcal X)}
\left(\int f\,dq-\int f\,dP-\frac14\int f^2\,dP\right).
\]
For TV, use the supremum of $|\int f\,dq-\int f\,dP|$ over
continuous $0\leq f\leq1$. Each ball contains its center, is compact,
and has a closed graph.

Let $z_n\to z$ and
$q\in\mathcal U_D(P_0(\cdot\mid z);\rho_D)$.
For $W_1$ and TV, changing the center changes the distance by $o(1)$.
For KL, the divergence changes by at most
$\|p_0(\cdot\mid z_n)-p_0(\cdot\mid z)\|_\infty/p_-$.
For $\chi^2$, the Lebesgue density of $q$ gives
\[
\chi^2(q\Vert P_0(\cdot\mid z_n))
\leq \rho_{\chi^2}
+\frac{1+\rho_{\chi^2}}{p_-}
\|p_0(\cdot\mid z_n)-p_0(\cdot\mid z)\|_\infty.
\]
Since $\rho_D>0$, convexity in the first argument makes
$(1-\epsilon)q+\epsilon P_0(\cdot\mid z_n)$ feasible for each fixed
$\epsilon\in(0,1)$ and all sufficiently large $n$.
A diagonal choice $\epsilon\downarrow0$ establishes lower hemicontinuity.
Berge's theorem gives continuity of the minimum and a nonempty compact
argmin with closed graph; measurable selection gives the Borel kernel.
\end{proof}

\begin{lemma}
\label{prop:rsfac-bellman}
Under Assumption~\ref{ass:cdm-regularity}, let $\pi$ be a weakly
continuous policy, meaning that $x\mapsto\int f(a)\pi(da\mid x)$ is
continuous for every continuous $f$ on $\mathcal A$.
Then $\mathcal T_D^\pi$ and $\mathcal T_D$ preserve $C(\mathcal X)$,
are monotone $\gamma$-contractions, and satisfy
\[
\mathcal T_D^\pi(v+c)=\mathcal T_D^\pi v+\gamma c,
\qquad
\mathcal T_D(v+c)=\mathcal T_Dv+\gamma c.
\]
Their fixed points satisfy $0\le V_D^\pi\le V_D^\star\le B$ and
\[
J_D(\pi)=\int V_D^\pi\,d\nu,
\qquad
\sup_\pi J_D(\pi)=\int V_D^\star\,d\nu.
\]
In the second identity, the supremum may include all policies
depending on the observed history, and an optimal Borel stationary
policy exists.
\end{lemma}
\begin{proof}
Lemma~\ref{lem:conditional-continuity} and weak continuity of $\pi$
give preservation of $C(\mathcal X)$. The inequality
\[
\left|\inf_{q\in\mathcal U_D(x,a)}\int v\,dq
-\inf_{q\in\mathcal U_D(x,a)}\int w\,dq\right|
\leq\|v-w\|_\infty
\]
gives contraction; monotonicity and the translation identities follow from integration.
Iteration from zero gives the bounds in $[0,B]$.
Since $\mathcal T_D^\pi V_D^\star\leq V_D^\star$, iteration also gives
$V_D^\pi\leq V_D^\star$.

For every $P\in\mathfrak P_D$,
$\mathcal T^{\pi,P}V_D^\pi\geq V_D^\pi$, so the ordinary policy value
is at least $V_D^\pi$. The Borel kernel minimizing
$\int V_D^\pi\,dq$ at every $(x,a)$ satisfies
$\mathcal T^{\pi,P}V_D^\pi=V_D^\pi$ and attains equality.

A Borel selector maximizing the continuous action value defined
by $V_D^\star$ gives a stationary policy satisfying
$r(X_t,A_t)+\gamma\mathbb E[V_D^\star(X_{t+1})\mid X_t,A_t]
\geq V_D^\star(X_t)$ under every $P\in\mathfrak P_D$.
Discounted telescoping gives return at least
$\int V_D^\star\,d\nu$.
Conversely, the Borel kernel minimizing $\int V_D^\star\,dq$
at every $(x,a)$ satisfies
\[
r(x,a)+\gamma\int V_D^\star(y)P(dy\mid x,a)\leq V_D^\star(x)
\quad\text{for every }x,a.
\]
Telescoping under this single kernel bounds every history-dependent
policy's return by $\int V_D^\star\,d\nu$, proving the second identity and attainment of the optimum.
\end{proof}

\begin{lemma}
\label{aux:approximate-robust-pi}
Let $\pi,\pi'$ be weakly continuous policies and $b\geq0$.
If $\mathcal T_D^{\pi'}V_D^\pi\geq\mathcal T_DV_D^\pi-b$ on
$\mathcal X$, then
\[
V_D^{\pi'}\geq\mathcal T_DV_D^\pi-\frac b{1-\gamma},\qquad
\|V_D^\star-V_D^{\pi'}\|_\infty
\leq\gamma\|V_D^\star-V_D^\pi\|_\infty+\frac b{1-\gamma}.
\]
\end{lemma}
\begin{proof}
Lemma~\ref{prop:rsfac-bellman} gives monotonicity, translation by
constants, contraction, and $\mathcal T_DV_D^\pi\geq V_D^\pi$.
Iterating the assumed inequality gives
$V_D^{\pi'}\geq V_D^\pi-b/(1-\gamma)$. Substitution into its
Bellman equation gives the first claim; contraction gives the second.
\end{proof}

\begin{lemma}\label{aux:uniform-residual-control}
Suppose Assumption~\ref{ass:cdm-regularity} holds, and let
$\pi_k$ be weakly continuous. Use the specified ReLU DNN classes
with fixed $L_0$ as in Lemma~\ref{thm:final-critic-rate}, and take
$\widehat Q_k\in\mathcal Q$. If
$\mu_k(dz)\geq p_-\epsilon\,dz$ almost surely, then
\[
\mathbb E\|\widehat Q_k-Q_k\|_\infty
\lesssim\frac{\epsilon^{-1/(d+2)}}{1-\gamma}
\left(\mathbb E\|\widehat Q_k-\mathcal B_k\widehat Q_k
\|_{L^2(\mu_k)}^2\right)^{1/(d+2)}.
\]
\end{lemma}
\begin{proof}
For every realization, the residual
$\widehat Q_k-\mathcal B_k\widehat Q_k$ is bounded in absolute value by $B$ and is
$2L_0$-Lipschitz by Lemma~\ref{aux:compatible-relu-classes}.
If its maximum norm is zero, Bellman contraction gives the claim.
Otherwise, its absolute value remains at least half its maximum
on the intersection of the cube with a maximum-norm ball centered
at a maximizer of the absolute residual, with radius
$
\frac{\|\widehat Q_k-\mathcal B_k\widehat Q_k\|_\infty}
{2(B+2L_0)}\leq\frac12.
$
This intersection has volume at least the radius to the power $d$.
The density floor therefore gives the pathwise inequality
\[
\epsilon\|\widehat Q_k-\mathcal B_k\widehat Q_k\|_\infty^{d+2}
\lesssim
\|\widehat Q_k-\mathcal B_k\widehat Q_k\|_{L^2(\mu_k)}^2.
\]
Since $\mathcal B_kQ_k=Q_k$, the triangle inequality and Bellman
contraction give
\[
(1-\gamma)\|\widehat Q_k-Q_k\|_\infty
\leq\|\widehat Q_k-\mathcal B_k\widehat Q_k\|_\infty.
\]
Combine these inequalities and apply Jensen's inequality with
exponent $1/(d+2)$ to obtain the stated expectation bound.
\end{proof}

\begin{lemma}
\label{prop:weighted-actor-greedy}
Under the conditions for \eqref{eq:actor-general-resolution}, suppose
additionally that $\widehat Q_k$ is $L_0$-Lipschitz in maximum
norm. Use the method's context and action augmentations at deployment,
together with uniform exploration, each with parameter $\epsilon$.
Then
\begin{align*}
&\mathbb E\sup_x\left[
\max_a\widehat Q_k(x,a)
-\int\widehat Q_k(x,a)\pi_{k+1}(da\mid x)\right]\lesssim
\lambda\log\frac{e}{\lambda\epsilon}+\epsilon
+C_{\alpha,\epsilon,B,\lambda}\epsilon^{-d_{\mathcal X}/2}
N^{-\alpha}(\log N)^{19/2}.
\end{align*}
The next policy has density
between $\epsilon$ and $\epsilon+C_{d_{\mathcal A}}\epsilon^{-d_{\mathcal A}}$ and
satisfies
\[
\TV\bigl(\pi_{k+1}(\cdot\mid x),
\pi_{k+1}(\cdot\mid x')\bigr)
\leq C_{d_{\mathcal X}}\epsilon^{-1}\|x-x'\|_\infty.
\]
\end{lemma}
\begin{proof}
Let $K_{r,q}(z,u)$ denote the density of $\mathsf F_q(u+rG)$ at $z$,
where $G\sim N(0,I_q)$.
At state $v$, let $a_\star$ maximize $\widehat Q_k(v,\cdot)$.
The cube intersection of maximum-norm radius $\lambda/(1+L_0)$
around $a_\star$ has volume at least
$[\lambda/(1+L_0)]^{d_{\mathcal A}}$; the KL divergence from its uniform law to
$\pi_k(\cdot\mid v)$ is at most
$\log(1/\epsilon)+d_{\mathcal A}\log((1+L_0)/\lambda)$.
The relative Gibbs variational formula gives
\[
\max_a\widehat Q_k(v,a)
-\int\widehat Q_k(v,a)\bar\pi_{k+1}(da\mid v)
\leq\lambda+\lambda\log(1/\epsilon)
+d_{\mathcal A}\lambda\log\frac{1+L_0}{\lambda}.
\]
Under $q_{k,\eta}(\cdot\mid u)$, the original context
$v=\mathsf F_{d_{\mathcal X}}(u+\eta G)$ is generated from Gaussian noise
reweighted by
$\int e^{\widehat Q_k(v,b)/\lambda}\pi_k(db\mid v)\in[1,e^{B/\lambda}]$.
The KL divergence of this reweighted noise law from the original Gaussian
is at most $B/\lambda$.
The Gaussian entropy inequality bounds the second moment of the
reweighted noise by $\lesssim B/\lambda+d_{\mathcal X}$, and hence
its expected context displacement by $\lesssim\eta\sqrt{1+B/\lambda}$.
Conditional on that context, the original action has law
$\bar\pi_{k+1}$ and its augmentation moves it by at most
$\eta\|G_a\|_2$. Lipschitz continuity of the critic therefore adds
$C L_0\eta\sqrt{1+B/\lambda}$; deployment context and action
augmentations add $C L_0\epsilon$.

Convexity and Cauchy--Schwarz give
\begin{align*}
&\sup_x\TV\left(
\int K_{\epsilon,d_{\mathcal X}}(x,u)\widehat\pi_{k+1}(\cdot\mid u)\,du,
\int K_{\epsilon,d_{\mathcal X}}(x,u)q_{k,\eta}(\cdot\mid u)\,du
\right)\\
&\quad\leq C_{d_{\mathcal X}}\epsilon^{-d_{\mathcal X}/2}
\left[\int\TV^2\bigl(\widehat\pi_{k+1}(\cdot\mid u),
q_{k,\eta}(\cdot\mid u)\bigr)\,du\right]^{1/2},
\end{align*}
because $\int K_{\epsilon,d_{\mathcal X}}^2(x,u)\,du
\leq\sup_u K_{\epsilon,d_{\mathcal X}}(x,u)
\leq C_{d_{\mathcal X}}\epsilon^{-d_{\mathcal X}}$.
The deployment action kernel contracts TV, and the expectation of a $[0,B]$-valued
function changes by at most $B$ times TV. Jensen's inequality and
\eqref{eq:actor-general-resolution}, followed by uniform exploration
at cost $B\epsilon$ and $\eta=\epsilon(1+B/\lambda)^{-1/2}$,
prove the greedy bound. The uniform component supplies the density
floor; the deployment action-kernel upper bound and the context-kernel
$L^1$ derivative bound give the upper density and TV Lipschitz bounds.
\end{proof}

\begin{lemma}\label{app:cdm-density}
For the fixed architecture, parameter bounds, positive cutoff, and
finite mesh in Section~\ref{form:actor-section}, each updated policy
admits a common-noise sampler with a deterministic state Lipschitz bound.
If $p_0\leq p_+$ and the offline query density is at least
$c_{\mathrm{off}}>0$, Assumption~\ref{ass:fixed-model-coverage} holds with
\[
C=\max\left\{1,\frac{p_+}{c_{\mathrm{off}}}
\max\left\{\sup_{x,a}\pi_0(a\mid x),
\epsilon+C_{d_{\mathcal A}}\epsilon^{-d_{\mathcal A}}\right\}\right\}.
\]
\end{lemma}
\begin{proof}
Fixed parameter bounds and a positive cutoff make each Euler update
globally Lipschitz in the noisy action and context, with deterministic
constants. Finite composition, folding, Gaussian augmentation, and a
shared Bernoulli selector for uniform exploration therefore give the
common-noise bound before oversampling. The positive-time state density
is bounded by $p_+$. At $k=0$, use the initial action-density upper bound;
for $k\geq1$, use Lemma~\ref{prop:weighted-actor-greedy}.
The offline density lower bound gives the stated coverage constant.
\end{proof}

Combining Lemmas~\ref{aux:approximate-robust-pi}--\ref{app:cdm-density}
with the critic bounds in Lemma~\ref{thm:final-critic-rate}, we now prove
Theorem~\ref{robust:global-finite-round}.

\begin{proof}[Proof of Theorem~\ref{robust:global-finite-round}]
Adding and subtracting $\widehat Q_k$ gives
\[
(\mathcal T_DV_D^{\pi_k})(x)
-(\mathcal T_D^{\pi_{k+1}}V_D^{\pi_k})(x)
\leq2\|\widehat Q_k-Q_k\|_\infty
+\max_a\widehat Q_k(x,a)-\int\widehat Q_k(x,a)\pi_{k+1}(da\mid x).
\]
Iterating Lemma~\ref{aux:approximate-robust-pi} and integrating
against $\nu$ give
\begin{align*}
&\mathbb E[\sup_\pi J_D(\pi)-J_D(\pi_K)]\\
&\leq B\gamma^K+
\frac1{1-\gamma}\sum_{k=0}^{K-1}\gamma^{K-1-k}
\left[2\mathbb E\|\widehat Q_k-Q_k\|_\infty
+\mathbb E\sup_x\left(\max_a\widehat Q_k(x,a)
-\int\widehat Q_k(x,a)\pi_{k+1}(da\mid x)\right)\right].
\end{align*}
Lemma~\ref{prop:weighted-actor-greedy} and the actor allocation
bound the second expectation by
$C[\lambda\log(e/(\lambda\epsilon))+\epsilon]$.
Apply Lemma~\ref{aux:uniform-residual-control} to
the expected residual bound in Lemma~\ref{cor:critic-fast-cone-rate} and the target-sampling and model bounds
proved for Lemma~\ref{thm:final-critic-rate}. The power is
$1/(d+2)$ for each residual component. The geometric sum is at most
$(1-\gamma)^{-1}$, and residual transfer contributes another
$(1-\gamma)^{-1}$.

The positive-time state density and deployed action-density bound give
\[
\frac{d\mu_k}{d\mu_{\mathrm{off}}}
\leq\frac{p_+}{c_{\mathrm{off}}}
\max\left\{\sup_{x,a}\pi_0(a\mid x),
\epsilon+C_{d_{\mathcal A}}\epsilon^{-d_{\mathcal A}}\right\}
\lesssim\epsilon^{-d_{\mathcal A}}.
\]
The calibration satisfies
\[
\epsilon^{-1/(d+2)}
\left[n^{-2/(d+2)}(\log n)^5\right]^{1/(d+2)}
=\epsilon,
\]
$\lambda\log(e/(\lambda\epsilon))\lesssim\epsilon$ and
$\gamma^K\leq\epsilon$. The target-sampling terms acquire the factor
$\epsilon^{-1/(d+2)}\leq n^{2/[(d+2)^2(d+3)]}$.
For W1, TV and the KL variance term,
\[
\epsilon^{-1/(d+2)}C^{1/(d+2)}
\lesssim n^{2(d_{\mathcal A}+1)/[(d+2)^2(d+3)]};
\]
for Pearson uncertainty,
\[
\epsilon^{-1/(d+2)}C^{1/[2(d+2)]}
\lesssim n^{(d_{\mathcal A}+2)/[(d+2)^2(d+3)]}.
\]
The $W_1$ and TV radius caps require
no coverage factor, and the pure KL cutoff ratios receive no additional
logarithmic factor.
\end{proof}

We apply Theorem~\ref{robust:global-finite-round} under the stated sample
allocations to prove Corollary~\ref{cor:robust-n-rate}.

\begin{proof}[Proof of Corollary~\ref{cor:robust-n-rate}]
For fixed positive $\rho$, the target-sampling terms for W1, TV,
and Pearson uncertainty are bounded by
$\widetilde{\mathcal O}(n^{2/[(d+2)^2(d+3)]}m^{-1/(d+2)})$.
Since $d\geq2$ and $m\geq n$, they are of order at most
$\widetilde{\mathcal O}(n^{-2/[(d+2)(d+3)]})$.
The stated polynomial allocation bounds the Pearson model term by
\[
n^{(d_{\mathcal A}+2)/[(d+2)^2(d+3)]}
n_x^{-\alpha/[(d_{\mathcal X}+d+2\alpha)(d+2)]}
\lesssim n^{-2/[(d+2)(d+3)]}.
\]
It also suffices for the W1 and TV model terms because
$2d+d_{\mathcal A}+6\geq d+d_{\mathcal A}+3$.
For KL, the exponential allocation gives
\[
n^{2/[(d+2)^2(d+3)]}
\left(\frac\rho{\rho+\log m}\right)^{2/(d+2)}
\lesssim n^{-2/[(d+2)(d+3)]},
\]
and the same bound holds with $m$ replaced by $n_x$. Its remaining
sampling and model terms, including their sample polylogarithms,
decay faster than every power of $n$. Theorem~\ref{robust:global-finite-round}
then gives the claim.
\end{proof}

\section{Auxiliary Results}\label{app:auxiliary-results}\label{app:cdm-import}
The reference bound below is used in both the nominal-transition analysis of
Lemma~\ref{thm:final-critic-rate} and the weighted actor analysis of
Lemma~\ref{thm:final-actor-rate}. We then apply it to the offline transition estimate.
\begin{lemma}[Convergence rate of conditional diffusion models]
\label{aux:conditional-diffusion}
Let $Z\in[0,1]^{d_{\mathcal Z}}$, and let $Y_0\mid Z=z$ have
conditional density $p_0(\cdot\mid z)$ on $[0,1]^{d_{\mathcal Y}}$.
Suppose the density, boundary, and conditioning regularity conditions
of Assumption~\ref{ass:cdm-regularity} hold in these dimensions. From $m$ i.i.d
pairs $(Y_0,Z)$, fit the unweighted reference score class at sufficiently
large resolution $N$, using the architecture, clipping, auxiliary
perturbations, and time truncation parameters of Section~\ref{sec:theory}, and the
reverse mesh specified in the proof below.
If $m\gtrsim N^{d_{\mathcal Y}+d_{\mathcal Z}+2\alpha}$, the resulting
folded conditional density estimate $\widehat p_0$ satisfies
\[
\mathbb E\int
\TV^2\bigl(\widehat p_0(\cdot\mid z),p_0(\cdot\mid z)\bigr)
\mathbb P_Z(dz)
\lesssim N^{-2\alpha}(\log N)^{19},
\]
where $\mathbb P_Z$ is the law of $Z$. The reference scores satisfy
$\|s(t,\cdot,\cdot)\|_\infty\lesssim
\sqrt{\log N}/\sqrt{1-e^{-t}}$.
\end{lemma}
\begin{proof}
The approximation and empirical-process bounds of
\citet[Lemmas~4.2--4.3]{Chen2026} apply at fixed resolution $N$.
Multiplying the diffusion times and mesh steps in that work by four
and composing each reference score's time input with $t\mapsto t/4$
gives the normalization in \eqref{eq:cdm-forward}.
This affine input change preserves the network architecture and the stated rates.
Lemma~4.3 of \citet{Chen2026} bounds the auxiliary-perturbation contribution by
$\lesssim N^{-2\alpha}(\log N)^{18}$.
The response-sample contribution is bounded by
\[
\frac{C N^{d_{\mathcal Y}+d_{\mathcal Z}}(\log N)^{14}
[(\log N)^4+\log m]}m
\lesssim N^{-2\alpha}(\log N)^{18},
\]
by $m\geq N^{d_{\mathcal Y}+d_{\mathcal Z}+2\alpha}$ and
$\log m/m\lesssim N^{-(d_{\mathcal Y}+d_{\mathcal Z}+2\alpha)}\log N$.
With the auxiliary budget
and time truncation parameters chosen from $N$, the same proof as
\citet[Theorem~4.1]{Chen2026} controls the remaining score error.
For the unprojected reference sampler, choose a deterministic grid with
maximum step satisfying
\[
\Delta\lesssim
\frac{N^{-2\alpha}}
{(\log N)^{21}\left[d_{\mathcal Y}(1+d_{\mathcal Y})M^{2(J+1)}W^{2J}
+N^{8\alpha+4\alpha/(\alpha\wedge1)}\right]}.
\]
This is the mesh condition in Remark~4.3 of \citet{Chen2026}, with its
network Lipschitz bound written explicitly and the factor-four time
change absorbed into the constant. It also meets that paper's
Lemma~4.4 step restriction. Its Theorem~4.2 and Lemmas~4.4--4.6 then give
the conditional TV bound, including initialization, truncation, and
finite sampling. Folding contracts TV.
\end{proof}

\paragraph{Nominal transition estimation.}
Under Assumption~\ref{ass:cdm-regularity} and the CDM sampling choices in Section~\ref{sec:theory},
Lemma~\ref{aux:conditional-diffusion}, with response
$X'\in[0,1]^{d_{\mathcal X}}$, conditioning variable
$(X,A)\in[0,1]^d$, design $\mu_{\mathrm{off}}$, and the prescribed
auxiliary samples and finite sampler, applies with
$N\asymp n_x^{1/(d_{\mathcal X}+d+2\alpha)}$ and gives
\begin{equation}
\mathbb E\int
\TV^2\!\left(\widehat P(\cdot\mid x,a),P_0(\cdot\mid x,a)\right)
\mu_{\mathrm{off}}(dx,da)
\lesssim n_x^{-2\alpha/(d_{\mathcal X}+d+2\alpha)}(\log n_x)^{19}.
\label{eq:conditional-model-rate}
\end{equation}
For every fitted history,
Assumption~\ref{ass:fixed-model-coverage} gives
\[
\int\TV^2\!\left(\widehat P(\cdot\mid x,a),P_0(\cdot\mid x,a)\right)
\mu_k(dx,da)
\leq C\int
\TV^2\!\left(\widehat P(\cdot\mid x,a),P_0(\cdot\mid x,a)\right)
\mu_{\mathrm{off}}(dx,da).
\]
Taking unconditional expectations transfers
\eqref{eq:conditional-model-rate} to the adaptive design $\mu_k$,
with multiplicative factor $C$.

\bibliographystyle{spbasic}
\bibliography{references}
\end{document}